\documentclass[sigconf]{acmart}
\usepackage{amsthm,amsmath}
\usepackage{graphicx}
\usepackage{subfigure}
\usepackage{mathrsfs}
\usepackage{algorithm}
\usepackage{algorithmicx}  
\usepackage{algpseudocode} 
 
\AtBeginDocument{%
  \providecommand\BibTeX{{%
    \normalfont B\kern-0.5em{\scshape i\kern-0.25em b}\kern-0.8em\TeX}}}

\copyrightyear{2021}
\acmYear{2021}
\setcopyright{acmcopyright}
\acmConference[MM '21]{Proceedings of the 29th ACM International Conference on Multimedia}{October 20--24, 2021}{Virtual Event, China} 
\acmBooktitle{Proceedings of the 29th ACM International Conference on Multimedia (MM '21), October 20--24, 2021, Virtual Event, China}
\acmPrice{15.00}
\acmDOI{10.1145/3474085.3475186}
\acmISBN{978-1-4503-8651-7/21/10}

\begin{document}
\fancyhead{}



\title{DSP: Dual Soft-Paste for Unsupervised Domain Adaptive Semantic Segmentation}

\author{Li Gao}
\authornote{This work was done during Li Gao's internship at JD Explore Academy.}
\email{gaoli1218@whu.edu.cn}
\affiliation{%
  \institution{Wuhan University}
  \country{China}
}

\author{Jing Zhang}
\email{jing.zhang1@sydney.edu.au}
\affiliation{%
  \institution{The University of Sydney}
  \country{Australia}}

\author{Lefei Zhang}
\authornote{Corresponding author.}
\email{zhanglefei@whu.edu.cn}
\affiliation{%
  \institution{Wuhan University}
  \country{China}
}

\author{Dacheng Tao}
\email{taodacheng@jd.com}
\affiliation{%
 \institution{JD Explore Academy}
 \country{China}}

\renewcommand{\shortauthors}{Li Gao, et al.}

\begin{abstract}

Unsupervised domain adaptation (UDA) for semantic segmentation aims to adapt a segmentation model trained on the labeled source domain to the unlabeled target domain. Existing methods try to learn domain invariant features while suffering from large domain gaps that make it difficult to correctly align discrepant features, especially in the initial training phase. To address this issue, we propose a novel Dual Soft-Paste (DSP) method in this paper. Specifically, DSP selects some classes from a source domain image using a long-tail class first sampling strategy and softly pastes the corresponding image patch on both the source and target training images with a fusion weight. 
Technically, we adopt the mean teacher framework for domain adaptation, where the pasted source and target images go through the student network while the original target image goes through the teacher network. Output-level alignment is carried out by aligning the probability maps of the target fused image from both networks using a weighted cross-entropy loss. In addition, feature-level alignment is carried out by aligning the feature maps of the source and target images from student network using a weighted maximum mean discrepancy loss. DSP facilitates the model learning domain-invariant features from the intermediate domains, leading to faster convergence and better performance. Experiments on two challenging benchmarks demonstrate the superiority of DSP over state-of-the-art methods. Code is available at \url{https://github.com/GaoLii/DSP}.

\end{abstract} 



\maketitle

\section{Introduction}
As one of the fundamental tasks in computer vision, semantic segmentation can be used as a preliminary step for many multimedia applications \cite{ye2020textfusenet,zhang2020empowering,chen2019progressive,vitae}, including image/video captioning, image-to-image translation, video content analysis. 
Training a well-performed deep semantic segmentation model usually requires a large amount of pixel-level labeled data, which is indeed very laborious and expensive to manually annotate. Alternatively, since it is much easier to generate synthetic images with dense pixel labels, $e.g.$, via a 3D game engine, there are a lot of works focusing on training the segmentation model using synthetic labeled images. However, due to the appearance discrepancy between synthetic images and real images, which is also known as the domain shift, the model trained on synthetic images usually generalizes poorly on real images.

\begin{figure}[t]
    \centering
    \includegraphics[width=\linewidth]{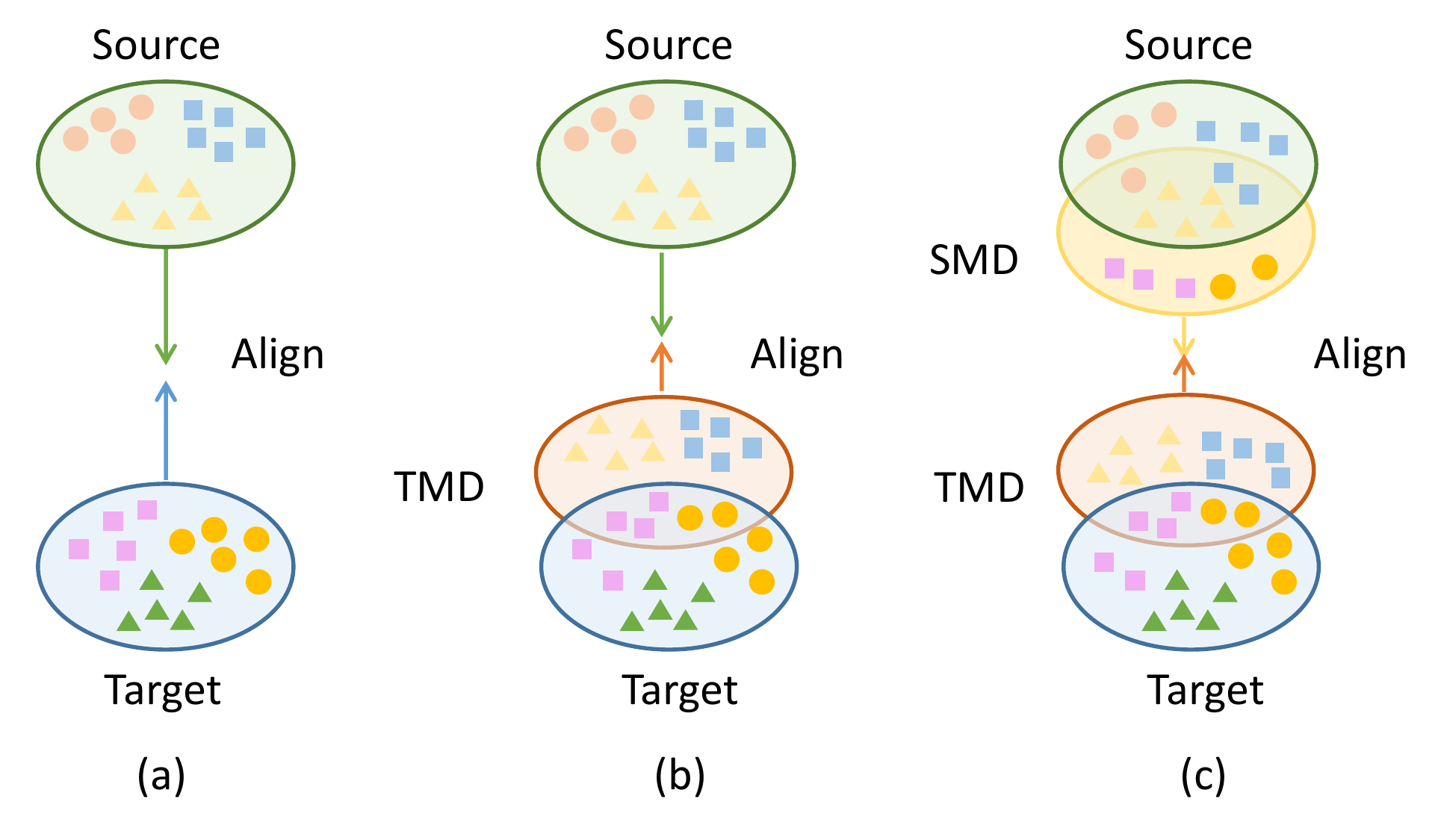}
    \caption{Illustration of different domain alignment paradigms. (a) Alignment on the original domains. (b) Alignment on the source domain and a mixed target domain, $e.g.$, DACS \cite{dacs}. (c) The proposed alignment method on two intermediate mixed domains by DSP. ``SMD'' and ``TMD'' refer to the source-mixed domain and target-mixed domain obtained by pasting source domain patches on the source and target domain images, respectively.
    }
    \label{main}
\end{figure}

To address this issue, unsupervised domain adaptation (UDA) methods have been proposed to mitigate the domain shift between the source and target domains as shown in Figure~\ref{main}(a). In the context of semantic segmentation, there are three categories of methods which perform domain adaptation at different levels, such as input level \cite{dacs,gl}, feature level \cite{cag,CLAN_pami,proda}, and output level \cite{fcns,crdoco}. Input-level UDA methods aim to perform statistical matching at the input level to achieve uniformity in the visual appearance of the input images from different domains, $e.g.$, style transfer \cite{cycada}. Feature-level UDA methods aim to align the distribution of latent features (usually embedded by CNNs) in both domains to extract domain-invariant features, $e.g.$, Maximum Mean Discrepancy (MMD) \cite{mmd}, adversarial learning \cite{gan}. Besides, since the predicted probability maps are in low-dimension and highly structured, it is effective to perform alignment of the probability maps from different domains, $i.e.$, output-level UDA \cite{crdoco}.

Recently, the mean teacher framework \cite{meanteacher} has been used by some methods for unsupervised domain adaptive semantic segmentation \cite{crdoco,TGCF}. These kinds of UDA methods perform output-level alignment by employing the consistency constraint on the target predictions from the student model and the teacher model, respectively. Though effective, they suffer from training instability and slow convergence due to inaccurate predictions on the unlabeled target domain, especially in the initial training phase. To address this issue, DACS \cite{dacs} proposes to paste part of a source domain image onto the unlabeled target domain image, which leads to certain parts of the pseudo-labeled map always being injected with the ground truth semantic map, ensuring the accuracy of the target prediction. Although DACS benefits from the intermediate mixed target domain as shown in Figure~\ref{main}(b), which helps to pull the target domain closer to the source domain, its performance is still limited due to the large gap between the two domains, incomplete structure, and inconsistent spatial layout issue by hard-paste, as well as the class imbalance issue.

In this paper, we go a step further and propose a novel paste method named dual soft-paste (DSP) for unsupervised domain adaptive semantic segmentation, which can create two new intermediate domains to facilitate aligning both source and target domains effectively, as shown in Figure \ref{main}. Specifically, DSP adopts a long-tail class first sampling strategy to select the candidate classes from a source domain template image and paste the corresponding image patch on both the source and target training images in a soft weighted-sum manner. It creates two new intermediate domains of fused images, which are used to perform domain alignment under the mean teacher framework. Technically, the pasted source and target images go through the student network while the original target image goes through the teacher network. Output-level alignment is carried out by aligning the probability maps of the fused target image from both networks using a weighted cross-entropy loss. Feature-level alignment is also performed by aligning the feature maps of the fused source and target image from student networks using a weighted maximum mean discrepancy loss. The dual paste strategy guarantees that the same patch is shared by both domain images serving as an intermediate to bridge both domains and the soft-paste strategy preserves the original domain information by maintaining its structure layout, complete objects, as well as appearance styles. Consequently, DSP facilitates the model learning domain-invariant features from the intermediate domains, leading to faster convergence and better performance. 

The contributions of this work can be summarized as follows:
\begin{itemize}
    \item We propose a novel Dual Soft-Paste (DSP) method to create intermediate domains and facilitate domain alignment for semantic segmentation. DSP adopts a long-tail class first sampling strategy, which alleviates the class imbalance issue and shows its effectiveness in improving the performance.
    
    \item Based on the mean teacher framework, we propose a new UDA model by performing both feature-level and output-level alignment on the intermediate domains, which benefits from the softly pasted patches with ground truth labels. 
    
    \item Extensive experiments on two challenging UDA semantic segmentation tasks, $i.e.$, GTA5 to Cityscapes and SYNTHIA to Cityscapes, clearly demonstrate the superiority of the proposed model over state-of-the-art methods.
\end{itemize}

\section{Related Work}

\begin{figure*}[htbp]
    \centering
    \includegraphics[width=\linewidth]{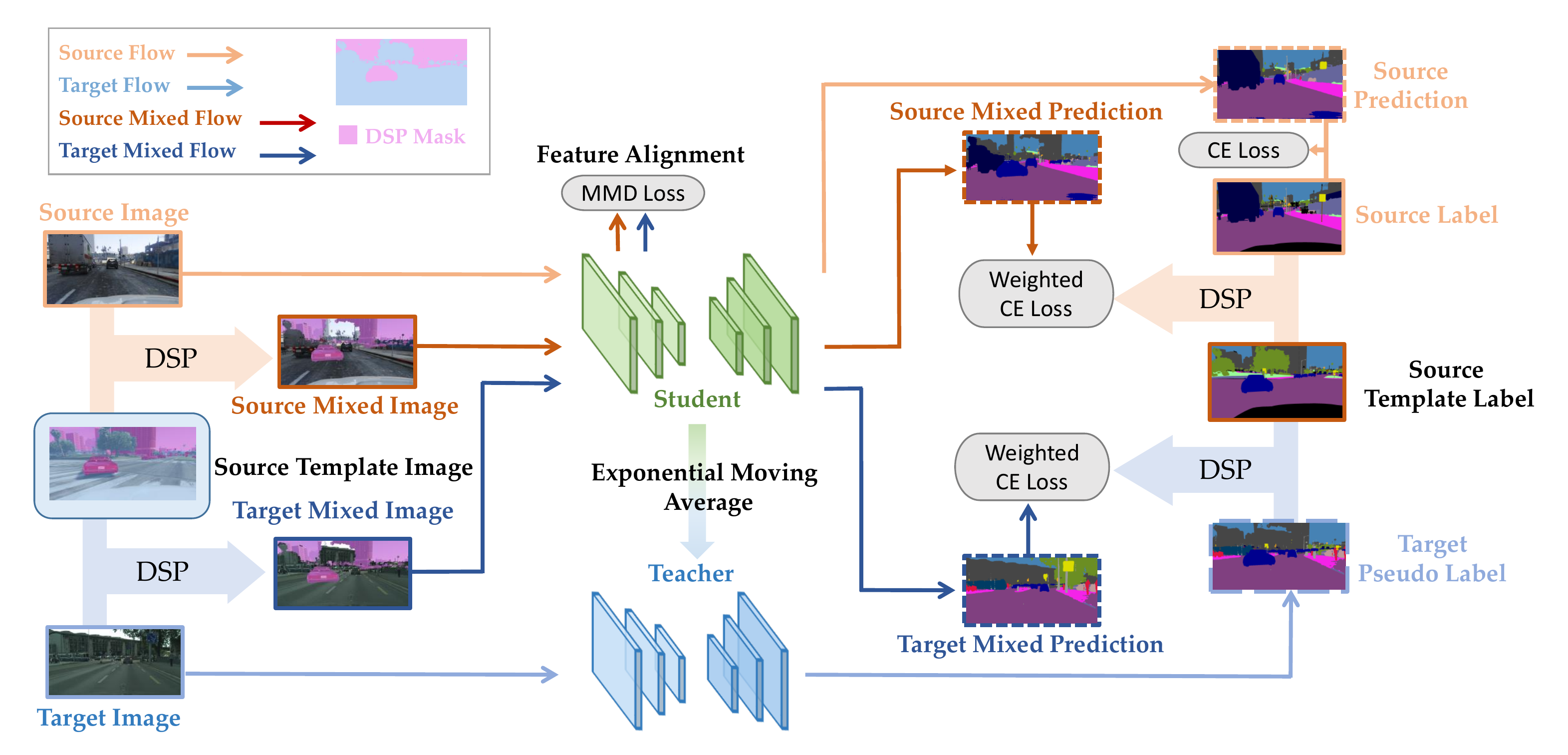}
    \caption{Overview of the proposed DSP model based on the mean teacher framework. The source image, source mixed image, and target mixed image are fed into the student network, while the target image is fed into the teacher network. DSP-induced output-level alignment and feature-level alignment are used to perform domain adaptation.}
    \label{framework}
\end{figure*}

\subsection{Mean Teacher-based Methods}
Since our work is built upon the mean teacher framework, we briefly review related methods in this section. For other UDA methods for semantic segmentation, such as style transfer and adversarial learning, we recommend the excellent survey \cite{review}. Mean teacher is a widely used framework in the field of semi-supervised learning, which is based on the simple idea that under the supervision of labeled data, unlabeled data should produce consistent predictions under different perturbations. It consists of two models, a student model and a teacher model, where the teacher model is an exponential moving average (EMA) of the student model. The teacher model transfers the learned knowledge to the student \cite{meanteacher} by aligning the two domains at the output level with a consistency regularization.

SEANET \cite{SEANET} firstly introduced the mean teacher framework for unsupervised domain adaptive semantic segmentation, which adopted an attention mechanism to generate attention-aware features to guide the calculation of consistency loss in the target domain. Zhou et al. \cite{uacr} proposed an uncertainty-aware consistency regularization method by exploiting the latent uncertainty information of the target samples. 
Recently, DACS \cite{dacs} proposed to paste source image patches onto the target domain images to create a mixed domain, where the labels of the pasted patches can be used for supervised learning and the prediction consistency between the student model and teacher model in the mean teacher framework is also exploited. Although DACS has achieved promising results, it still suffers from several problems, including the large gap between two domains, the incomplete structure and inconsistent spatial layout problem by hard-paste, as well as the class imbalance problem, resulting in limited performance. 

Different from DACS, we propose a novel dual soft-paste method to solve the aforementioned problems. First, our dual paste strategy can create two intermediate domains by pasting same source image patches on both the source and target images, which indeed serves as a bridge to reduce the gap between the two domains. Second, our soft-paste strategy can preserve the original domain information by keeping its structure layout, complete objects, as well as appearance styles. Third, we perform both feature-level and output-level alignment to learn domain-invariant features from the intermediate domains, leading to faster convergence and better performance.

\subsection{Copy-and-Paste Strategies}
There is a wide spectrum of work to improve the performance of deep models by using copy-and-paste methods for data augmentation in the supervised training setting. For example, CutMix \cite{cutmix} cut and pasted patches among training images where the labels are also mixed to the area of the patches. Remez et al. learned object masks by cutting-and-pasting with adversarial learning \cite{lscp}. However, all these methods adopted the hard-paste strategy, which loses the original layout and semantic information of the original image when creating new images. MixUp \cite{mixup} trained the network on convex combinations of image pairs and their labels with a mixing weight to address the aforementioned issue. Dwibedi et al. proposed to automatically cut object instances and paste them on random backgrounds to make detectors ignore these artifacts during training and generate data that gives competitive performance on real data \cite{cpl}. 
FMix \cite{fmix} proposed to use random binary masks obtained by applying a threshold to low frequency images sampled from fourier space. These random masks can take on a wide range of shapes and can be generated for use with one, two, and three dimensional data. In this paper, we also explore the soft-paste idea but specifically tailor it to the unsupervised domain adaptive semantic segmentation setting by handling the class imbalance issue and reducing the domain gap. 

\subsection{Self-Training}
Self-training is a widely used strategy for semi/unsupervised learning by generating pseudo labels of the unlabeled data. For the semantic segmentation task, CBST \cite{cbst} proposed an iterative self-training method that alternatively generated pseudo labels on target data via latent variable loss minimization and retrained the model using these labels. DAST \cite{DAST} presented a discriminator attention-based self-training method to adaptively improve the decision boundary of the model for the target domain. IAST \cite{IAST} developed a pseudo-label generation strategy, which uses an instance adaptive selector and a region-guided regularization to smooth the pseudo-label region and sharpen the non-pseudo-label region. Zheng et al. explicitly estimated the prediction uncertainty during training to rectify the pseudo label learning \cite{rpll}. MetaCorrection \cite{MetaCorrection} proposed to model the noise distribution of pseudo labels in the target domain to advance domain-aware meta learning. We also leverage the self-training idea but implement it together with the proposed DSP method under the mean teacher framework, where the target pseudo labels are generated by the teacher model, mixed with the pasted source patch labels, and updated during training. 

\section{Method}
\subsection{Preliminaries}
Denoting the source domain by $S$, it contains images $X_S$ and pixel-level labels $Y_S$, while the target domain $T$ only contains unlabeled images $X_T$. The goal of UDA-based semantic segmentation is training a model on $\{X_S, Y_S, X_T\}$ that can predict accurate semantic labels for $X_T$. To this end, we proposed a novel DSP model under the mean teacher framework as illustrated in Figure \ref{framework}. It has two segmentation networks, $i.e.$, a student network $f_{\theta}$ with learnable parameters $\theta$ and a teacher network $f_{\theta'}$ with parameters $\theta'$ calculated by the exponential moving average (EMA) of $\theta'$.

\renewcommand{\algorithmicrequire}{\textbf{Input:}}  
\renewcommand{\algorithmicensure}{\textbf{Output:}} 
\begin{algorithm}[t]  
  \caption{The proposed dual soft-paste algorithm}  
  \begin{algorithmic}[1]  
    \Require  
       Source template image $x_p$ and its label $y_p$, source image $x_s$, target image $x_t$, pre-defined long-tail dataset $D$, opacity $\beta$;
    \Ensure  
      The DSP mask $M$, mixed source image $x_{ps}$, mixed target image $x_{pt}$;  
    \State $S_{class} \gets$ set of classes appearing in $y_p$ ;  
    \State $c \gets$ randomly select $|S_{class}|/2$ classes from $S_{class}$;  
    
    \For{each $i, j$}  
      \State 
$$ M(i, j)=\left\{
\begin{aligned}
1 ,\quad & if \ y_p(i,j) \in c \\
0 ,\quad & else
\end{aligned}
\right.
$$ 
    \EndFor 
    \State $c_{tail}, y_{tail} \gets$ randomly select source images from $D$; 
     \For{each $i, j$} 
        \If{$y_{tail}(i,j) \in c_{tail}$} $M(i,j) = 1 $
        \EndIf
    \EndFor \\
    $x_{ps}, x_{pt} \gets$ calculate mixed images by Eq. \ref{eq2} and Eq. \ref{eq3}
    \label{code:recentEnd}
    \State \Return $M$, $x_{ps}$, $x_{pt}$;
  \end{algorithmic}  
\end{algorithm}

\subsection{Dual Soft-Paste}
\subsubsection{Long-Tail Class First Sampling}
Given the images from source domain $S$, we first calculate the frequency distribution of their classes as $\{p_1, p_2, ..., p_c\}$, $i.e.$,
\begin{equation}
    p_{i} = \frac{\sum_{j=1}^{N} c_{ij}}{N},
\end{equation}
where $c$ represents the number of categories, $c_{ij}$ indicates whether $j-th$ source image contains class $c_i$, and $N$ denotes the total number of images in $S$. Then, we choose the least frequent $K$ categories as the long-tail categories and record those images containing these classes as a dataset $D$ to facilitate the subsequent sampling process. In this paper, we set $K$ to 5 for the GTA5 dataset, including rider, bus, train, motorbike, bike. And $K = 4$ for the SYNTHIA dataset, including wall, light, bus, bike. During training, we randomly choose $k$ long-tail classes and select $k$ images from $D$, each of which contains at least one of the chosen long-tail classes. $k$ is set to 2 in this paper. A hyper-parameter study of $k$ is conducted in Section~\ref{subsec:ablationstudy}.

\subsubsection{The Algorithm of Dual Soft-Paste}
\begin{figure}[t]
    \centering
    \subfigure[Source Image]{
    \includegraphics[width=0.45\linewidth,height = 0.22\linewidth]{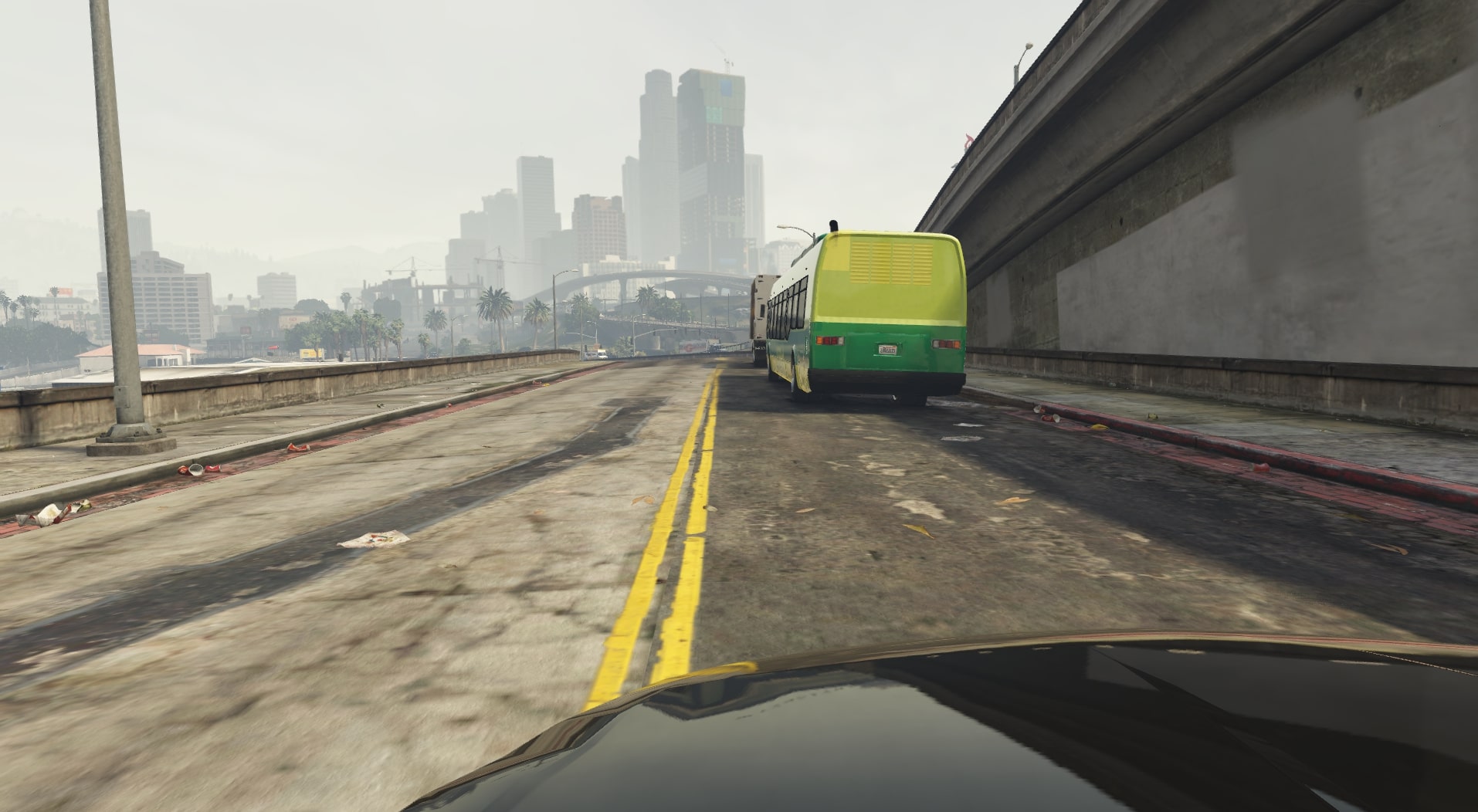}}
    \subfigure[Target Image]{
    \includegraphics[width=0.45\linewidth,height = 0.22\linewidth]{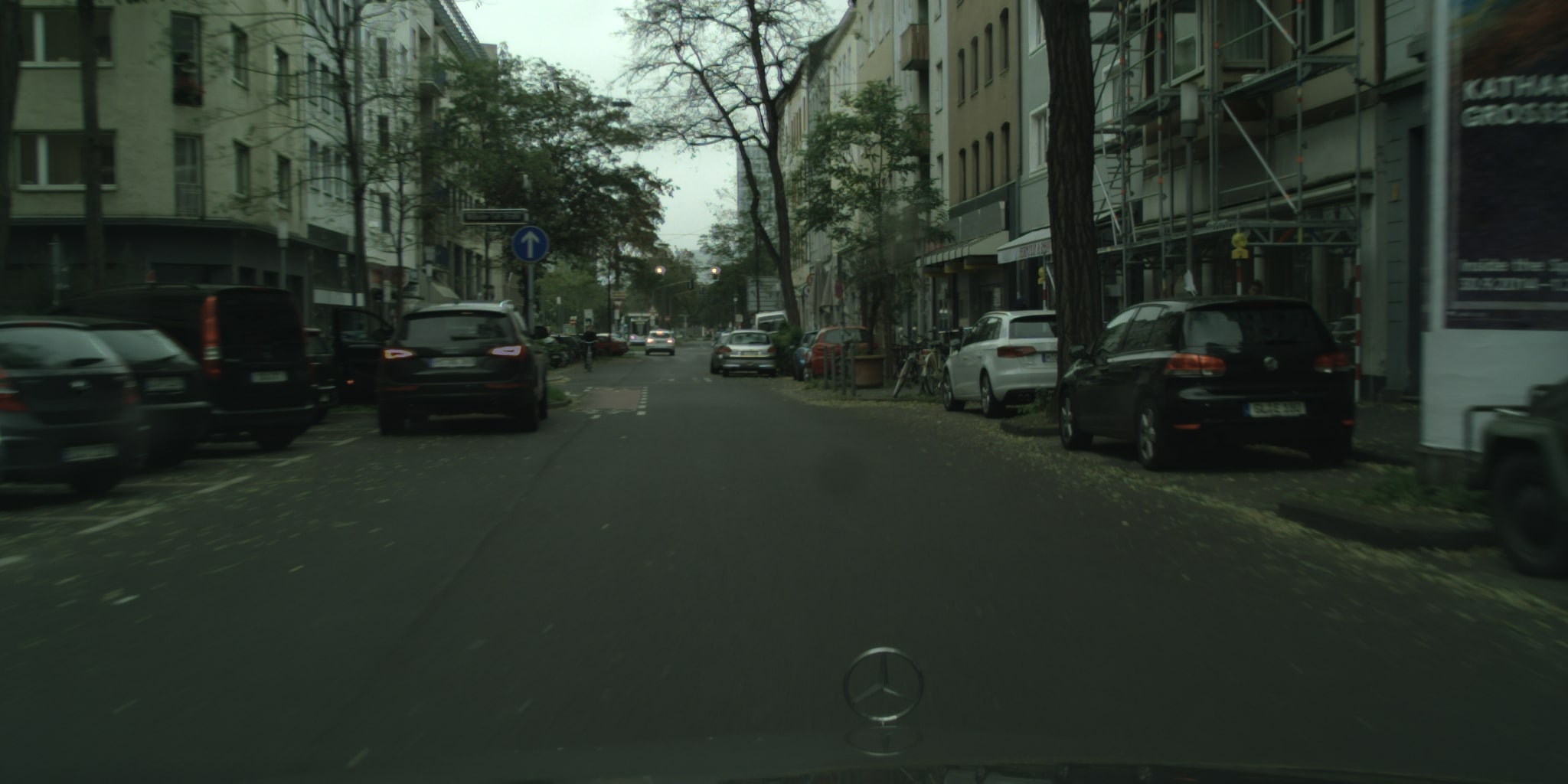}}
    
    \subfigure[Source Template Image]{
    \includegraphics[width=0.45\linewidth,height = 0.22\linewidth]{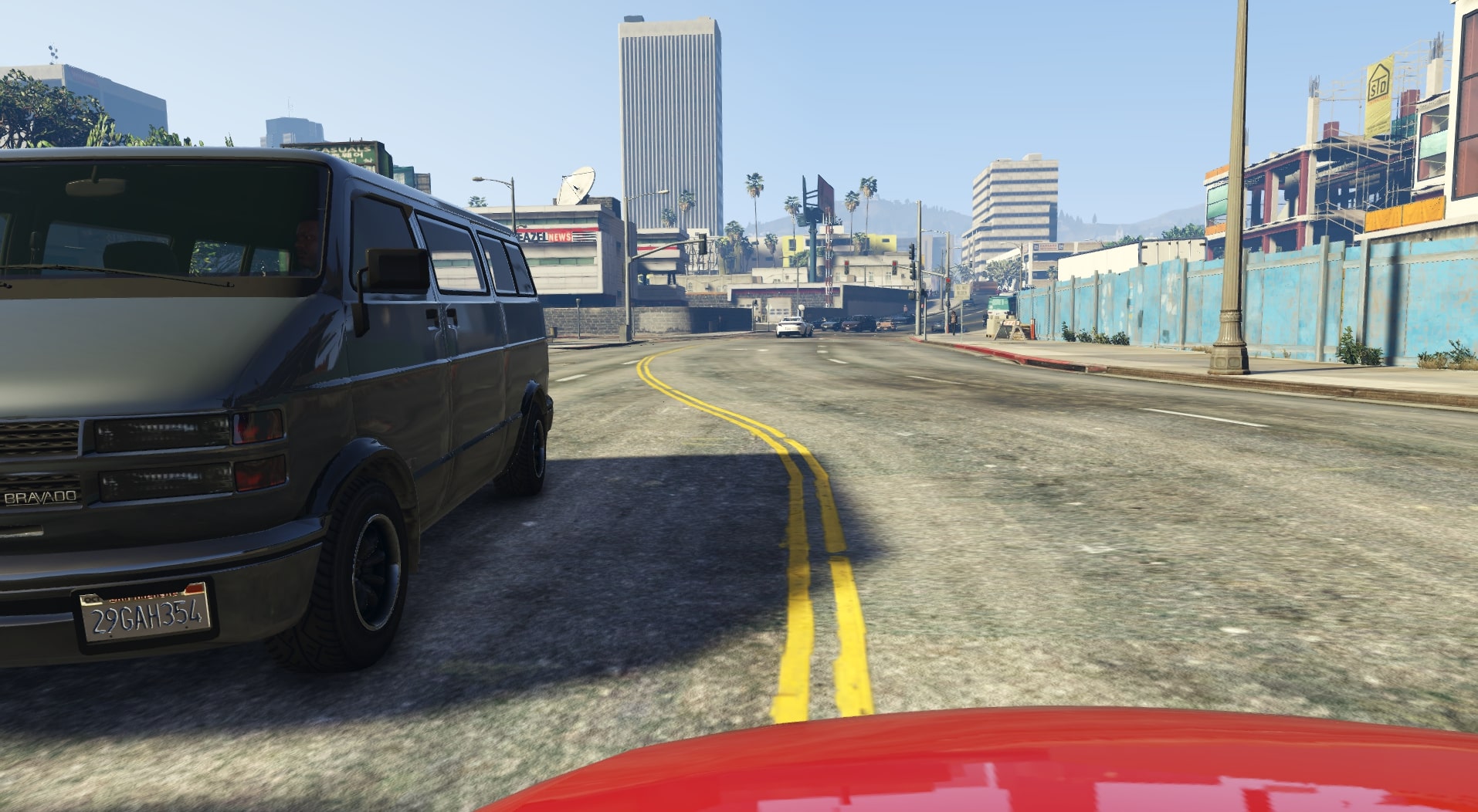}}
    \subfigure[DSP Mask]{
    \includegraphics[width=0.45\linewidth,height = 0.22\linewidth]{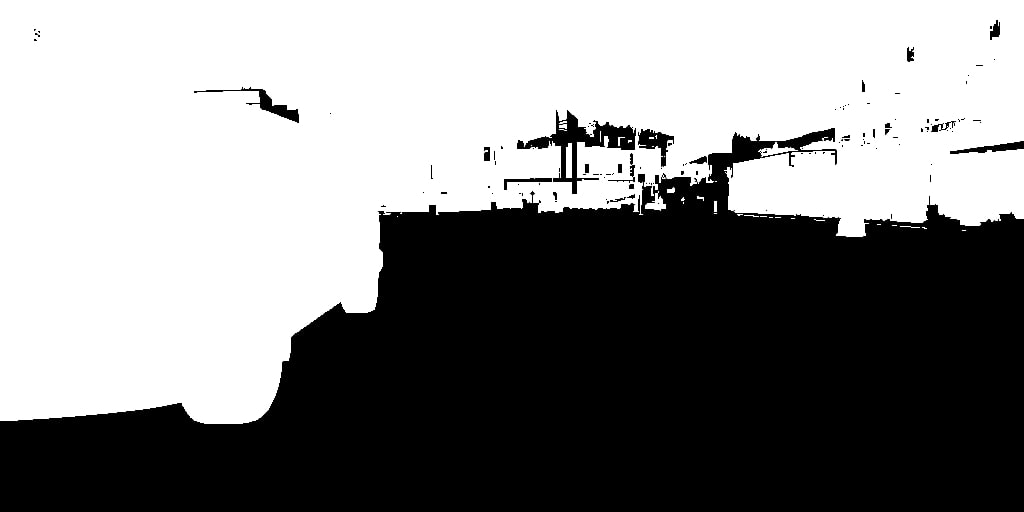}}
    
    \subfigure[Source Soft Mixed Image]{
    \includegraphics[width=0.45\linewidth,height = 0.22\linewidth]{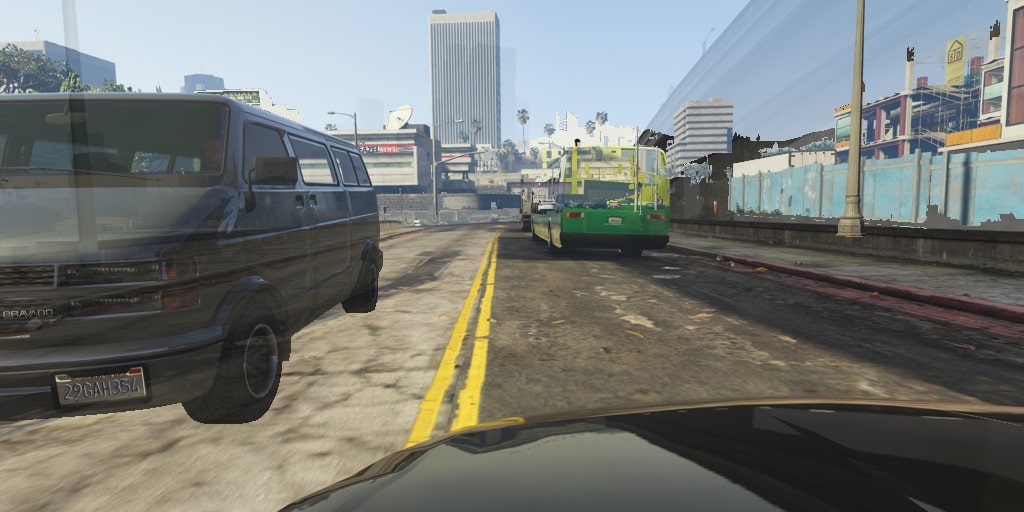}}
    \subfigure[Target Soft Mixed Image]{
    \includegraphics[width=0.45\linewidth,height = 0.22\linewidth]{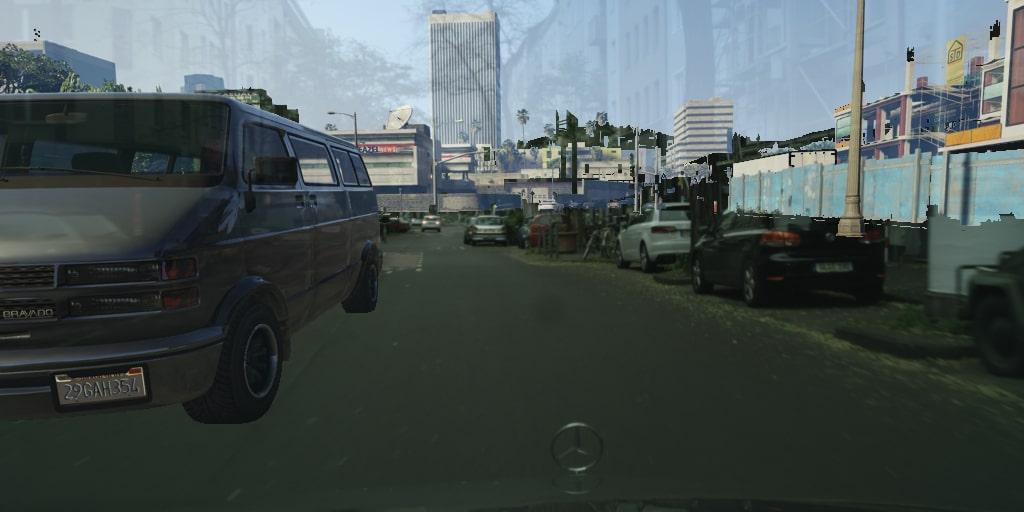}}
    \caption{Visual demonstration of the DSP algorithm. Note that we show the binary DSP mask instead of the soft one by integrating the opacity $\beta$ for better illustration. }
    \label{dsp_images}
\end{figure}

During training, we first randomly choose a source image from $S$ and select half of its classes and corresponding image patch as the candidate patch used for subsequent pasting. Then, we choose long-tail classes and candidate images as described above. Next, we merge the candidate patch with the patches of long-tail classes to form the final candidate patch. In this way, we can guarantee that the candidate patches have both frequent classes and long-tail classes. Then, we paste this patch on a source image and target image via a soft weighted-sum manner using an opacity parameter $\beta$. Specifically, given a source image $x_s$, a target image $x_t$, and the source template image $x_p$ with the corresponding binary mask $M$, the mixed source image $x_{ps}$ can be obtained by:
\begin{equation}
    x_{ps}  =  \beta M  \odot  x_p  +  (1 - \beta M) \odot  x_s.
    \label{eq2}
\end{equation}
Similarly, the mixed target image $x_{pt}$ can be obtained as follows:
\begin{equation}
    x_{pt} =  \beta M  \odot x_p  +  (1 - \beta M) \odot \ x_t.
    \label{eq3}
\end{equation}
For simplicity, we reuse $M$ to represent $\beta M$ by assigning the opacity value to those positive pixels in $M$. A hyper-parameter study of $\beta$ is conducted in Section~\ref{subsec:ablationstudy}, which is set to 0.8 by default. The algorithm of DSP is presented in Algorithm 1. In addition, we show a visual example of DSP in Figure~\ref{dsp_images}.

It is noteworthy that DSP has the following merits: First, it creates two mixed images that share an identical source template image patch at the same location, which can serve as a bridge to effectively reduce the domain gap between both domains. Second, it preserves the original domain information by keeping its structure layout, complete objects, as well as appearance style. Third, the ground truth labels of pasted source image patches can be leveraged for output-level alignment.

\begin{table*}[thb]
\normalsize
\centering
\resizebox{\textwidth}{!}{
\begin{tabular}{c|ccccccccccccccccccc|c}
\hline
\multicolumn{21}{c}{GTA5 $\to$ Cityscapes}\\
\hline
Method & 
\rotatebox{90}{road} & 
\rotatebox{90}{sidewalk} & 
\rotatebox{90}{building}&
\rotatebox{90}{wall}&
\rotatebox{90}{fence}&
\rotatebox{90}{pole}&
\rotatebox{90}{light}&
\rotatebox{90}{sign}&
\rotatebox{90}{vegetation}&
\rotatebox{90}{terrain}&
\rotatebox{90}{sky}&
\rotatebox{90}{person}&
\rotatebox{90}{rider}&
\rotatebox{90}{car}&
\rotatebox{90}{trunk}&
\rotatebox{90}{bus}&
\rotatebox{90}{train}&
\rotatebox{90}{motorbike}&
\rotatebox{90}{bike}&
mIoU\\
\hline\hline
Source Only&75.8&16.8&77.2&12.5&21.0&25.5&30.1&20.1&81.3&24.6&70.3&53.8&26.4&49.9&17.2&25.9&6.5&25.3&36.0&36.6\\
WeakSeg(ECCV20)\cite{WEEKSEGDA}&91.6&47.4&84.0&30.4&28.3&31.4&37.4&35.4&83.9&38.3&83.9&61.2&28.2&83.7&28.8&41.3&8.8&24.7&46.4&48.2\\
LSE(ECCV20)\cite{lse}&90.2&40.0&83.5&31.9&26.4&32.6&38.7&37.5&81.0&34.2&84.6&61.6&33.4&82.5&32.8&45.9&6.7&29.1&30.6&47.5\\
IAST(ECCV20)\cite{IAST}&94.1&58.8&85.4&\textbf{39.7}&29.2&25.1&43.1&34.2&84.8&34.6&88.7&62.7&30.3&87.6&42.3&50.3&24.7&35.2&40.2&52.2\\
CrCDA(ECCV20)\cite{CRCDA}&92.4&55.3&82.3&31.2&29.1&32.5&33.2&35.6&83.5&34.8&84.2&58.9&32.2&84.7&40.6&46.1&2.1&31.1&32.7&48.6\\
LTIR(CVPR20)\cite{ltir}&92.9&55.0&85.3&34.2&31.1&34.9&40.7&34.0&85.2&40.1&87.1&61.0&31.1&82.5&32.3&42.9&0.3&36.4&46.1&50.2\\
UIDA(CVPR20)\cite{UIDA}&90.6&37.1&82.6&30.1&19.1&29.5&32.4&20.6&85.7&40.5&79.7&58.7&31.1&86.3&31.5&48.3&0.0&30.2&35.8&46.3\\
PIT(CVPR20)\cite{PIT}&87.5&43.4&78.8&31.2&30.2&36.3&39.9&42.0&79.2&37.1&79.3&65.4&\textbf{37.5}&83.2&46.0&45.6&25.7&23.5&49.9&50.6\\
STAR(CVPR20)\cite{STAR}&88.4&27.9&80.8&27.3&25.6&26.9&31.6&20.8&83.5&34.1&76.6&60.5&27.2&84.2&32.9&38.2&1.0&30.2&31.2&43.6\\
ASA(TIP21)\cite{asa}&89.2&27.8&81.3&25.3&22.7&28.7&36.5&19.6&83.8&31.4&77.1&59.2&29.8&84.3&33.2&45.6&16.9&34.5&30.8&45.1\\
CLAN(TPAMI21)\cite{CLAN_pami}&88.7&35.5&80.3&27.5&25.0&29.3&36.4&28.1&84.5&37.0&76.6&58.4&29.7&81.2&38.8&40.9&5.6&32.9&28.8&45.5\\
DACS(WACV21)\cite{dacs}&89.9&39.7&\textbf{87.9}&\textbf{39.7}&\textbf{39.5}&\textbf{38.5}&46.4&\textbf{52.8}&\textbf{88.0}&\textbf{44.0}&88.8&\textbf{67.2}&35.8&84.5&45.7&50.2&0.0&27.3&34.0&52.1\\
RPLL(IJCV21)\cite{rpll}&90.4&31.2&85.1&36.9&25.6&37.5&\textbf{48.8}&48.5&85.3&34.8&81.1&64.4&36.8&86.3&34.9&52.2&1.7&29.0&44.6&50.3\\
DAST(AAAI21)\cite{DAST}&92.2&49.0&84.3&36.5&28.9&33.9&38.8&28.4&84.9&41.6&83.2&60.0&28.7&87.2&45.0&45.3&7.4&33.8&32.8&49.6\\
ConTrans(AAAI21)\cite{ct}&\textbf{95.3}&\textbf{65.1}&84.6&33.2&23.7&32.8&32.7&36.9&86.0&41.0&85.6&56.1&25.9&86.3&34.5&39.1&11.5&28.3&43.0&49.6\\
CIRN(AAAI21)\cite{gl}&91.5&48.7&85.2&33.1&26.0&32.3&33.8&34.6&85.1&43.6&86.9&62.2&28.5&84.6&37.9&47.6&0.0&35.0&36.0&49.1\\
MetaCorrect(CVPR21)\cite{MetaCorrection}&92.8&58.1&86.2&\textbf{39.7}&33.1&36.3&42.0&38.6&85.5&37.8&87.6&62.8&31.7&84.8&35.7&50.3&2.0&36.8&48.0&52.1\\
ESL(CVPR21)\cite{esl}&90.2&43.9&84.7&35.9&28.5&31.2&37.9&34.0&84.5&42.2&83.9&59.0&32.2&81.8&36.7&49.4&1.8&30.6&34.1&48.6\\
\textbf{Our DSP}&92.4&48.0&87.4&33.4&35.1&36.4&41.6&46.0&87.7&43.2&\textbf{89.8}&66.6&32.1&\textbf{89.9}&\textbf{57.0}&\textbf{56.1}&0.0&\textbf{44.1}&\textbf{57.8}&\textbf{55.0}\\
\hline
\end{tabular}}
\caption{Results of different domain adaptation methods for the GTA5 $\to$ Cityscapes task.}
\label{gta2city}
\end{table*}

\subsection{Mean Teacher-based Domain Adaptation}
\subsubsection{DSP-induced Output-level Alignment}

During training, the original source image ($x_s$), the mixed source and target image ($x_{ps}$ and $x_{pt}$) are fed into the student network $f_{\theta}$, while the original target image ($x_t$) is fed into the teacher network $f_{\theta'}$. While the parameters $\theta$ of the student network are optimized via gradient back-propagation, the parameters $\theta'_t$ of the teacher network at training step $t$ are updated using EMA as follows:
\begin{equation}
    \theta'_t = \alpha \cdot \theta'_{t-1} + (1 - \alpha) \cdot \theta_t,
\end{equation}
where $\alpha$ denotes the EMA decay coefficient.

\begin{figure}[t]
    \centering
    \includegraphics[width=\linewidth]{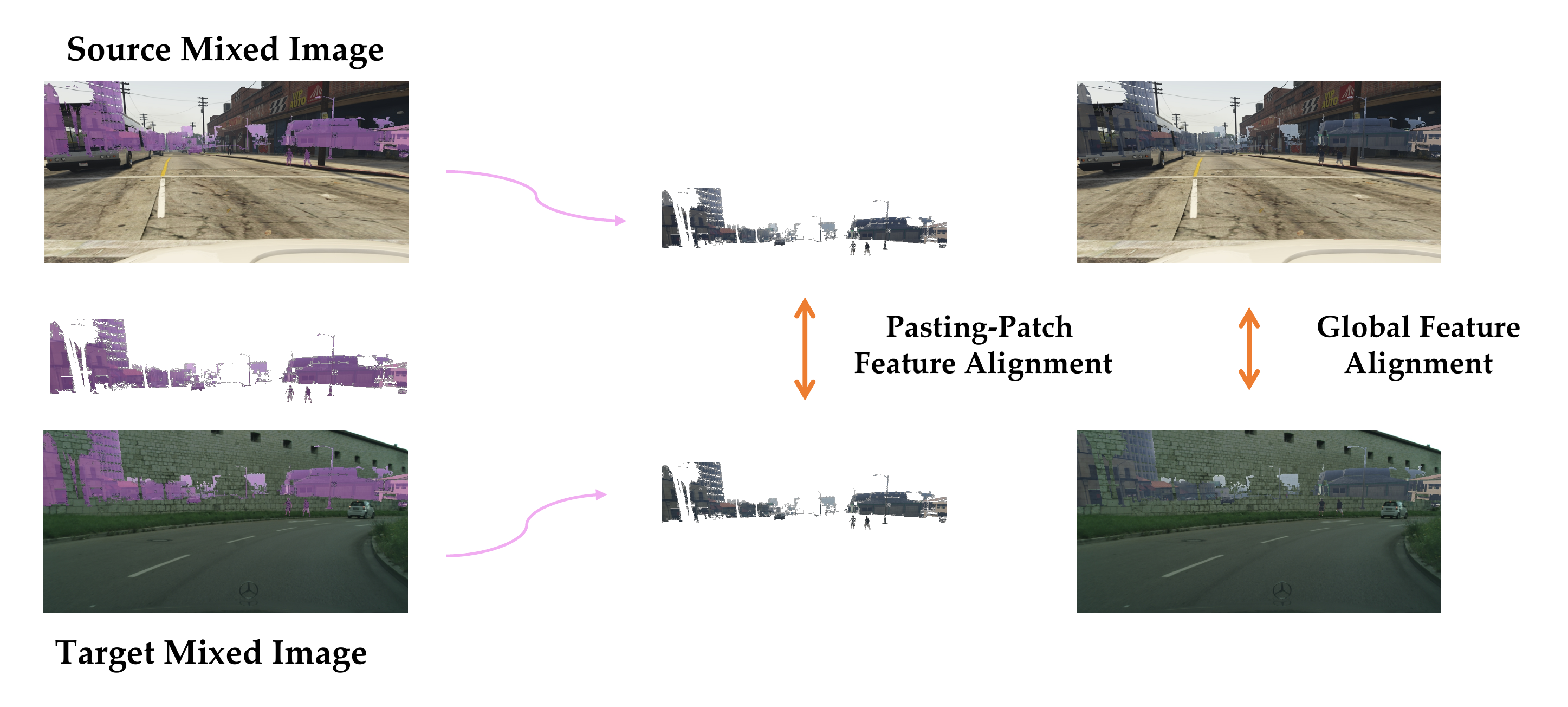}
    \caption{Illustration of DSP-induced feature alignment.}
    \label{mmd}
\end{figure}

After obtaining the predict semantic map $p_s$ of $x_s$, a cross-entropy based \textit{semantic segmentation loss} is used for training the network:
\begin{equation}
    \mathcal{L}_{seg} = -\sum_{i=1}^{H\times W}\sum_{c=1}^{C}y_{s}^{H\times W \times  C } log p_{s}^{H\times W \times  C },
\end{equation}
where $H, W, C$ represent the height and width of the image, and number of classes, respectively. $y_s$ denotes the ground truth semantic labels. 

Similarly, after obtaining the predict semantic map $p_{ps}$ of $x_{ps}$ from $f_{\theta}$, we use a weighted cross-entropy based \textit{soft semantic segmentation loss} to train the network, $i.e.$, 
\begin{equation}
\begin{split}
    \mathcal{L}_{seg\_soft} = &-\sum_{i=1}^{H\times W}\sum_{c=1}^{C}y_{p}^{H\times W \times  C } log p_{ps}^{H\times W \times  C } \odot M \\
    &- \sum_{i=1}^{H\times W}\sum_{c=1}^{C}y_{s}^{H\times W \times  C } log p_{ps}^{H\times W \times  C } \odot (1 - M),
    \end{split}
\end{equation}
where $y_p$ is the ground truth semantic label of the pasted source image patch. 

From the teacher model $f_{\theta'}$, the pseudo label of the original target image $x_t$ can be obtained as $\hat{y}_t = f_{\theta'}(x_{t})$. Meanwhile, we can obtain the predict semantic map $p_{pt}$ of $x_{pt}$ from $f_{\theta}$. Since these two models are assumed to produce a same prediction for a same image under different perturbations, so we adopt a \textit{prediction consistency loss} to train the network, $i.e.$,
\begin{equation}
\begin{split}
    \mathcal{L}_{cons} = &-\sum_{i=1}^{H\times W}\sum_{c=1}^{C}y_{p}^{H\times W \times  C } log p_{pt}^{H\times W \times  C } \odot M\\
    &-\sum_{i=1}^{H\times W}\sum_{c=1}^{C}\hat{y}_{t}^{H\times W \times  C } log p_{pt}^{H\times W \times  C } \odot (1 - M).
    \end{split}
\end{equation}

\subsubsection{DSP-induced Feature-level Alignment}

\begin{table*}[ht]
\centering
\resizebox{\textwidth}{!}{
\begin{tabular}{c|cccccccccccccccc|c|c}
\hline
\multicolumn{19}{c}{SYNTHIA $\to$ Cityscapes}\\
\hline
Method & 
\rotatebox{90}{road} & 
\rotatebox{90}{sidewalk} & 
\rotatebox{90}{building}&
\rotatebox{90}{wall*}&
\rotatebox{90}{fence*}&
\rotatebox{90}{pole*}&
\rotatebox{90}{light}&
\rotatebox{90}{sign}&
\rotatebox{90}{vegetation}&
\rotatebox{90}{sky}&
\rotatebox{90}{person}&
\rotatebox{90}{rider}&
\rotatebox{90}{car}&
\rotatebox{90}{bus}&
\rotatebox{90}{motorbike}&
\rotatebox{90}{bike}&
mIoU&
mIoU*\\
\hline\hline
Source Only&55.6&23.8&74.6&9.2&0.2&24.4&6.1&12.1&74.8&79.0&55.3&19.1&39.6&23.3&13.7&25.0&33.5&38.6\\
WeakSeg(ECCV20)\cite{WEEKSEGDA}&92.0&53.5&80.9&11.4&0.4&21.8&3.8&6.0&81.6&84.4&60.8&24.4&80.5&39.0&26.0&41.7&44.3&51.9\\
LSE(ECCV20)\cite{lse}&82.9&43.1&78.1&9.3&0.6&28.2&9.1&14.4&77.0&83.5&58.1&25.9&71.9&38.0&29.4&31.2&42.6&49.4\\
IAST(ECCV20)\cite{IAST}&81.9&41.5&\textbf{83.3}&17.7&\textbf{4.6}&32.3&30.9&28.8&83.4&85.0&65.5&30.8&\textbf{86.5}&38.2&\textbf{33.1}&52.7&49.8&57.0\\
CrCDA(ECCV20)\cite{CRCDA}&86.2&44.9&79.5&8.3&0.7&27.8&9.4&11.8&78.6&86.5&57.2&26.1&76.8&39.9&21.5&32.1&42.9&50.0\\
LTIR(CVPR20)\cite{ltir}&92.6&53.2&79.2&-&-&-&1.6&7.5&78.6&84.4&52.6&20.0&82.1&34.8&14.6&39.4&-&49.3\\
UIDA(CVPR20)\cite{UIDA}&84.3&37.7&79.5&5.3&0.4&24.9&9.2&8.4&80.0&84.1&57.2&23.0&78.0&38.1&20.3&36.5&41.7&48.9\\
PIT(CVPR20)\cite{PIT}&83.1&27.6&81.5&8.9&0.3&21.8&26.4&\textbf{33.8}&76.4&78.8&64.2&27.6&79.6&31.2&31.0&31.3&44.0&51.8\\
STAR(CVPR20)\cite{STAR}&82.6&36.2&81.1&-&-&-&12.2&8.7&78.4&82.2&59.0&22.5&76.3&33.6&11.9&40.8&-&48.1\\
ASA(TIP21)\cite{asa}&91.2&48.5&80.4&3.7&0.3&21.7&5.5&5.2&79.5&83.6&56.4&21.9&80.3&36.2&20.0&32.9&41.7&49.3\\
CLAN(TPAMI21)\cite{CLAN_pami}&82.7&37.2&81.5&-&-&-&17.1&13.1&81.2&83.3&55.5&22.1&76.6&30.1&23.5&30.7&-&48.8\\
DACS(WACV21)\cite{dacs}&80.6&25.1&81.9&\textbf{21.5}&2.9&\textbf{37.2}&22.7&24.0&83.7&\textbf{90.8}&\textbf{67.6}&\textbf{38.3}&82.9&38.9&28.5&47.6&48.3&54.8\\
RPLL(IJCV21)\cite{rpll}&87.6&41.9&83.1&14.7&1.7&36.2&31.3&19.9&81.6&80.6&63.0&21.8&86.2&40.7&23.6&53.1&47.9&54.9\\
DAST(AAAI21)\cite{DAST}&87.1&44.5&82.3&10.7&0.8&29.9&13.9&13.1&81.6&86.0&60.3&25.1&83.1&40.1&24.4&40.5&45.2&52.5\\
ConTrans(AAAI21)\cite{ct}&\textbf{93.3}&\textbf{54.0}&81.3&14.3&0.7&28.8&21.3&22.8&82.6&83.3&57.7&22.8&83.4&30.7&20.2&47.2&46.5&53.9\\
CIRN(AAAI21)\cite{gl}&85.8&40.4&80.4&4.7&1.8&30.8&16.4&18.6&80.7&80.4&55.2&26.3&83.9&43.8&18.6&34.3&43.9&51.1\\
MetaCorrect\cite{MetaCorrection}(CVPR21)&92.6&52.7&81.3&8.9&2.4&28.1&13.0&7.3&83.5&85.0&60.1&19.7&84.8&37.2&21.5&43.9&45.1&52.5\\
ESL(CVPR21)\cite{esl}&84.3&39.7&79.0&9.4&0.7&27.7&16.0&14.3&78.3&83.8&59.1&26.6&72.7&35.8&23.6&45.8&43.5&50.7\\
\textbf{Our DSP}&86.4&42.0&82.0&2.1&1.8&34.0&\textbf{31.6}&33.2&\textbf{87.2}&88.5&64.1&31.9&83.8&\textbf{65.4}&28.8&\textbf{54.0}&\textbf{51.0}&\textbf{59.9}\\
\hline
\end{tabular}}
\caption{Results of different domain adaptation methods for the SYNTHIA $\to$ Cityscapes task. mIoU* denotes the mean IoU of 13 classes, excluding the classes marked by the asterisk.}
\label{syn2city}
\end{table*}

Since the source mixed image $x_{ps}$ and target mixed image $x_{pt}$ have the same pasted source image patch, the features extracted in this region from $x_{ps}$ and $x_{pt}$ should be as similar as possible. To this end, we adopt Maximum Mean Discrepancy (MMD) \cite{mmd} to learn transferable features by minimizing the MMD of their kernel embeddings. This \textit{paste-patch feature alignment loss} can be formulated as:
\begin{equation}
    \mathcal{L}_{paste} = \left \| \mu({f_e(x_{ps}) \odot M})- \mu({f_e(x_{pt}) \odot M})\right \|_{\mathcal{H}}^{2},
\end{equation}
where $\mu(\cdot)$ denotes the kernel mean embedding, $f_e$ represents the feature extractor of the student model $f_{\theta}$ ($i.e.$, the network before the ASPP module), and $\mathcal{H}$ denotes the reproducing kernel Hilbert space (RKHS). Note that since the pasted patches in $x_{ps}$ and $x_{pt}$ have different context of source and target domain information, which may be embedded in the extracted features, the paste-patch alignment loss can reduce the domain gap implicitly.

In addition, we try to minimize the MMD of the image features of $x_{ps}$ and $x_{pt}$ to align the feature distributions of both domains. This \textit{global feature alignment loss} is: 
\begin{equation}
    \mathcal{L}_{global} = \left \| \mu(f_e(x_{ps}))- \mu(f_e(x_{pt}))\right \|_{\mathcal{H}}^{2}.
\end{equation}
An illustration of these two losses is shown in Figure \ref{mmd}. The overall training objective can be defined as:
\begin{equation}
    \mathcal{L} = \mathcal{L}_{seg} + \mathcal{L}_{seg\_soft} + \mathcal{L}_{cons}+ \lambda_{feature} (\mathcal{L}_{paste} + \mathcal{L}_{global}),
\end{equation}
where $\lambda_{feature}$ is a hyper-parameter to balance different losses. 

\section{Experiments}

\begin{figure}[tbhp]
    \centering
    \subfigure[]{
    \includegraphics[width=1\linewidth]{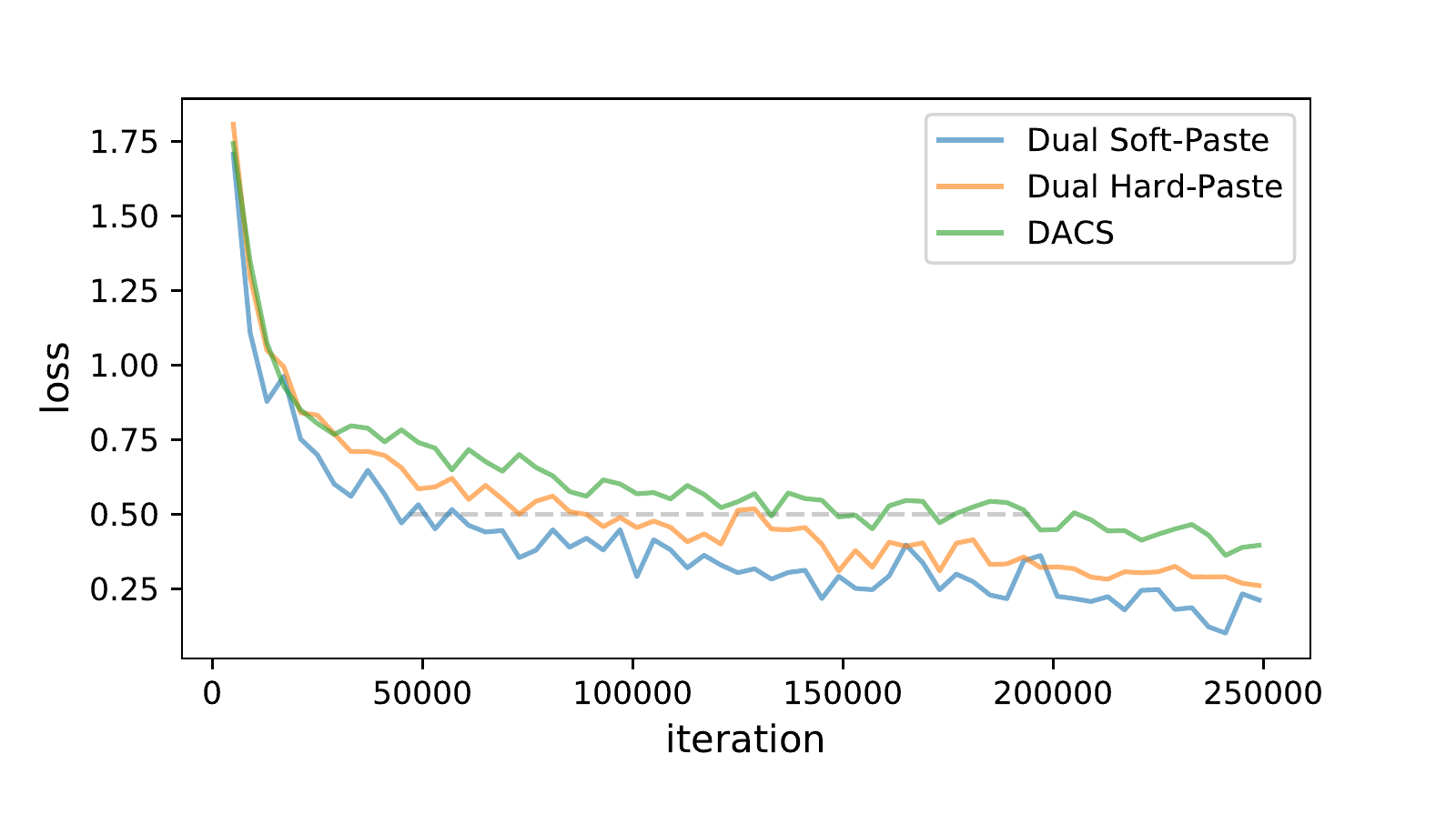}}
    \subfigure[]{
    \includegraphics[width=1\linewidth]{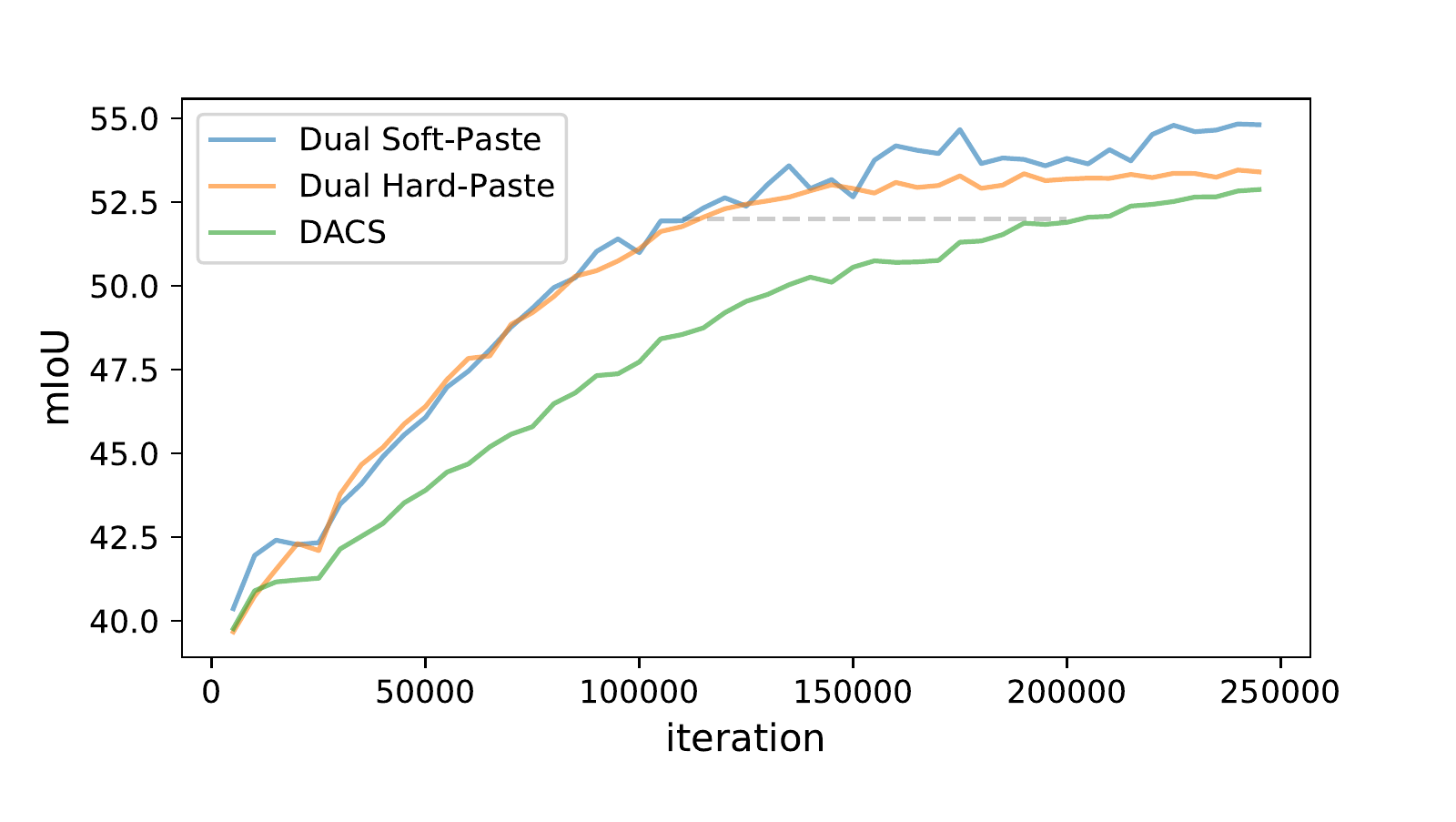}}
    \caption{Convergence analysis of DACS \cite{dacs}, Dual Hard-Paste (a variant of the proposed DSP based on hard-paste), and the proposed DSP on the GTA5 to Cityscapes setting. (a) Training losses. (b) mIoU results. DSP converges faster, $i.e.$, achieving better performance in the same training steps.}
    \label{loss_miou}
    
\end{figure}
\subsection{Datasets and Evaluation Metrics}
We evaluate the performance of the proposed method for two challenging UDA tasks: GTA5 \cite{GTA5} to Cityscapes \cite{Cityscapes} and SYNTHIA \cite{SYNTHIA} to Cityscapes. GTA5 is a synthetic dataset created using a photo-realistic open-world computer game engine. Dense pixel-level semantic annotations are provided for 24,966 urban landscape images with a resolution of $1,914 \times 1,052$. 19 common classes in the Cityscapes dataset are chosen in our experiments. SYNTHIA is another synthetic collection of 9,400 diverse urban images with a resolution of $1,280 \times 760$. We consider 16 common categories in the Cityscapes dataset for evaluation while the results on 13 common classes are also reported following a common practice. Cityscapes is a large-scale real-world urban scenes benchmark for semantic segmentation, which provides 5,000 densely annotated images with a resolution of $2,048 \times 1,024$. We use 2,975 unlabeled training images during training and 500 validation images for testing. In all experiments, we use the mIoU metric for evaluation. 

\subsection{Implementation Details}
The proposed model is implemented using PyTorch on a single NVIDIA Tesla V100 GPU with 16 GB memory. Following previous work, we adopt ResNet-101 \cite{resnet} pre-trained on ImageNet \cite{imagenet} and on MSCOCO \cite{coco} as the backbone network to extract features, and ASPP \cite{deeplabv2} is adopted to be the classifier to predict semantic maps. We use Stochastic Gradient Descent (SGD) with Nesterov acceleration as the optimizer, an initial learning rate of $2.5 \times 10^{-3}$ for the feature extractor, and an initial learning rate of $2.5 \times 10^{-4}$ for the classifier, which are then decreased based on a polynomial decay policy with an exponent of 0.9. Weight decay is set to $5 \times 10^{-4}$ and momentum is set to 0.9. During training, we resize images in Cityscapes, GTA5, and SYNTHIA to $1,024 \times 512$, $1,280 \times 720$, and $1,280 \times 760$, respectively, after which the input images are randomly cropped into patches with a size of 512 $\times$ 512. We also apply color jittering and Gaussian blurring for data augmentation. The model is trained for a total of 250,000 iterations with a batch size of 2. We set $\lambda_{feature}$ to 0.005, the EMA decay coefficient $\alpha$ to 0.99, opacity $\beta$ to 0.8, and the number of long-tail classes selected at each training iteration $k$ to 2.

\subsection{Comparison with State-of-the-art Methods}
In this section, we evaluate our DSP model on the two challenging UDA semantic segmentation tasks and compare it with several state-of-the-art methods.

\begin{figure*}[htbp]
\centering

\subfigure[Target Images]{
\begin{minipage}[t]{0.24\linewidth}
\centering
\includegraphics[width=1.65in,height=0.83in]{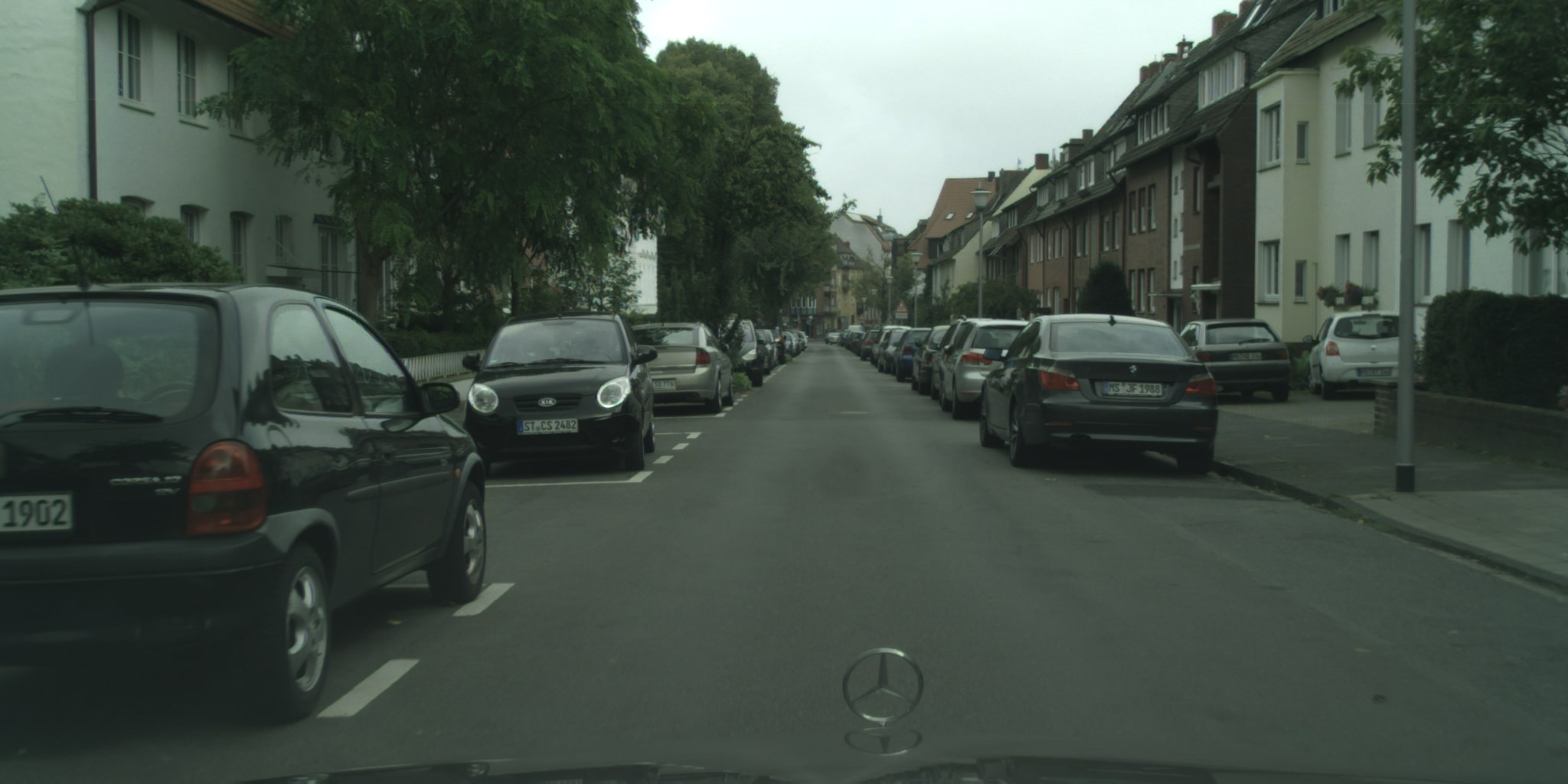}
\includegraphics[width=1.65in,height=0.83in]{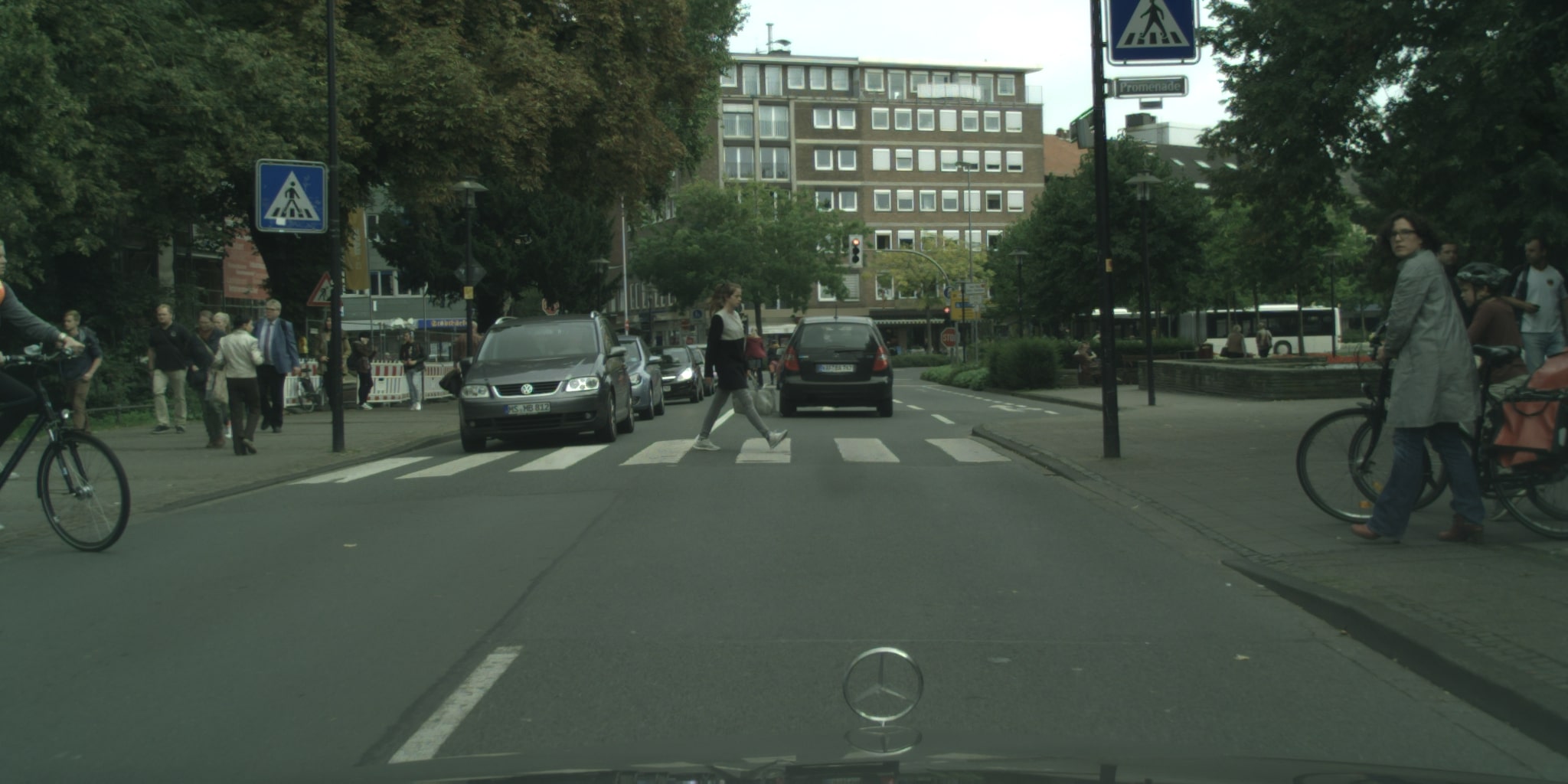}
\includegraphics[width=1.65in,height=0.83in]{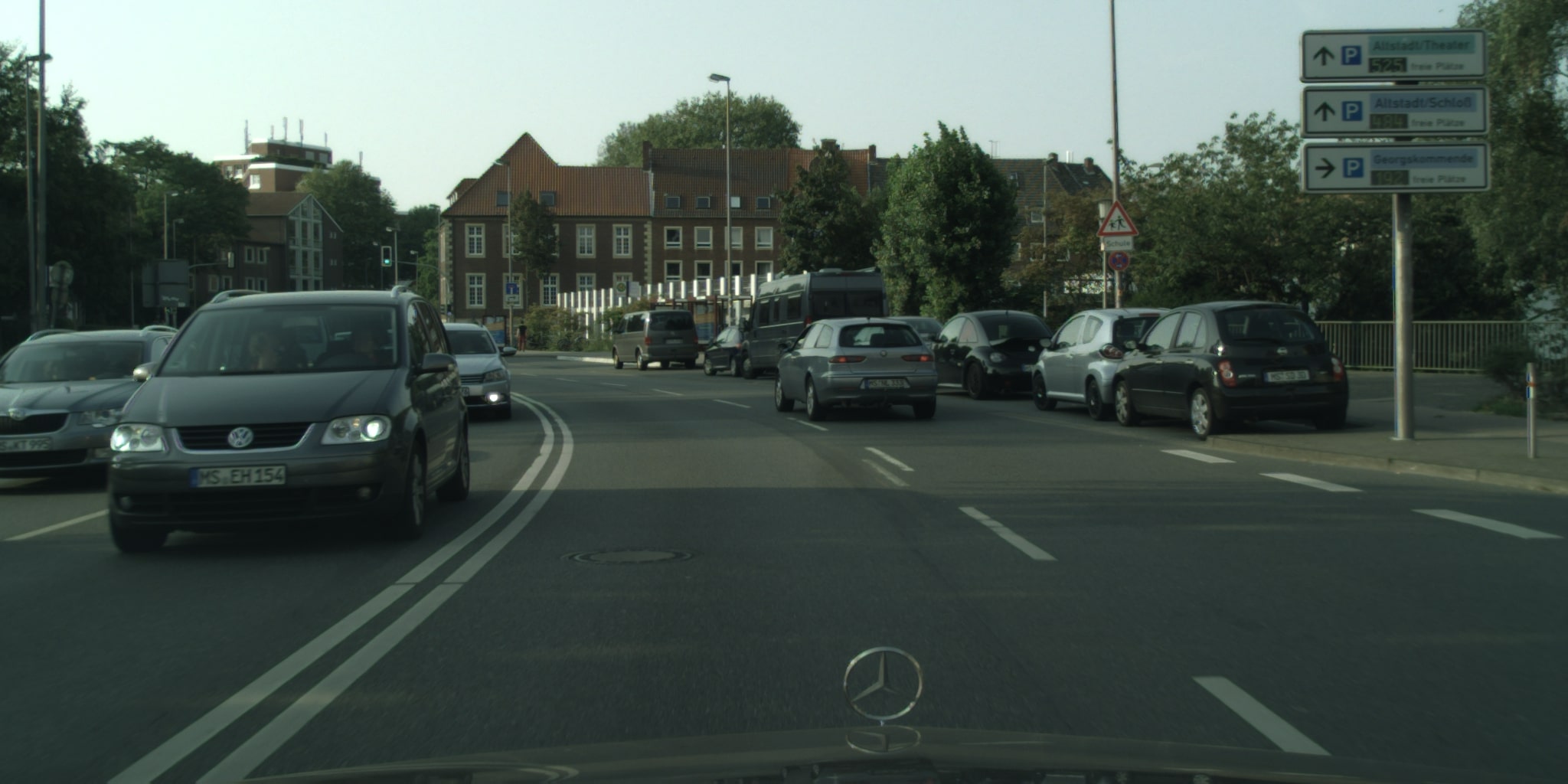}
\includegraphics[width=1.65in,height=0.83in]{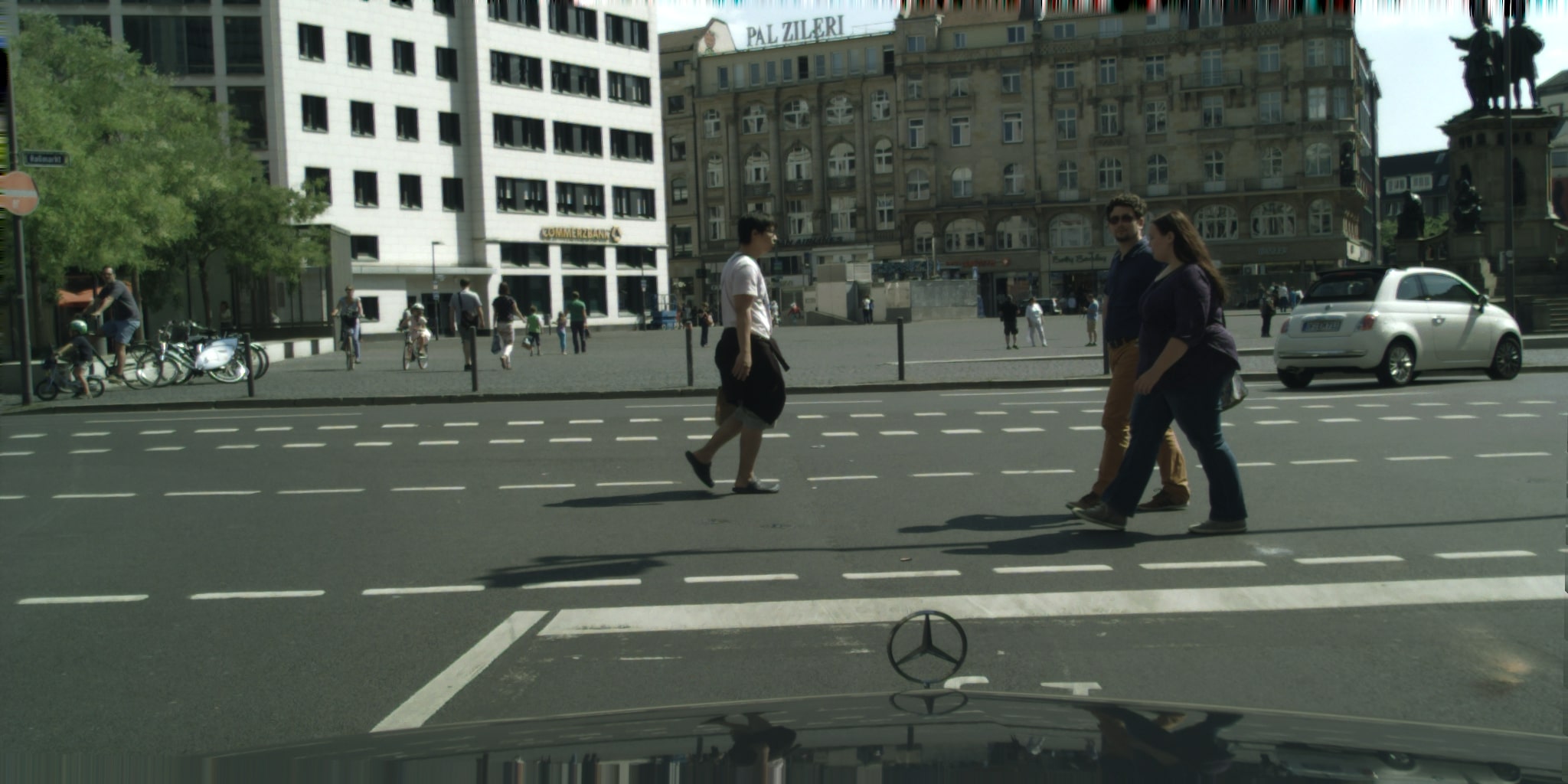}
\includegraphics[width=1.65in,height=0.83in]{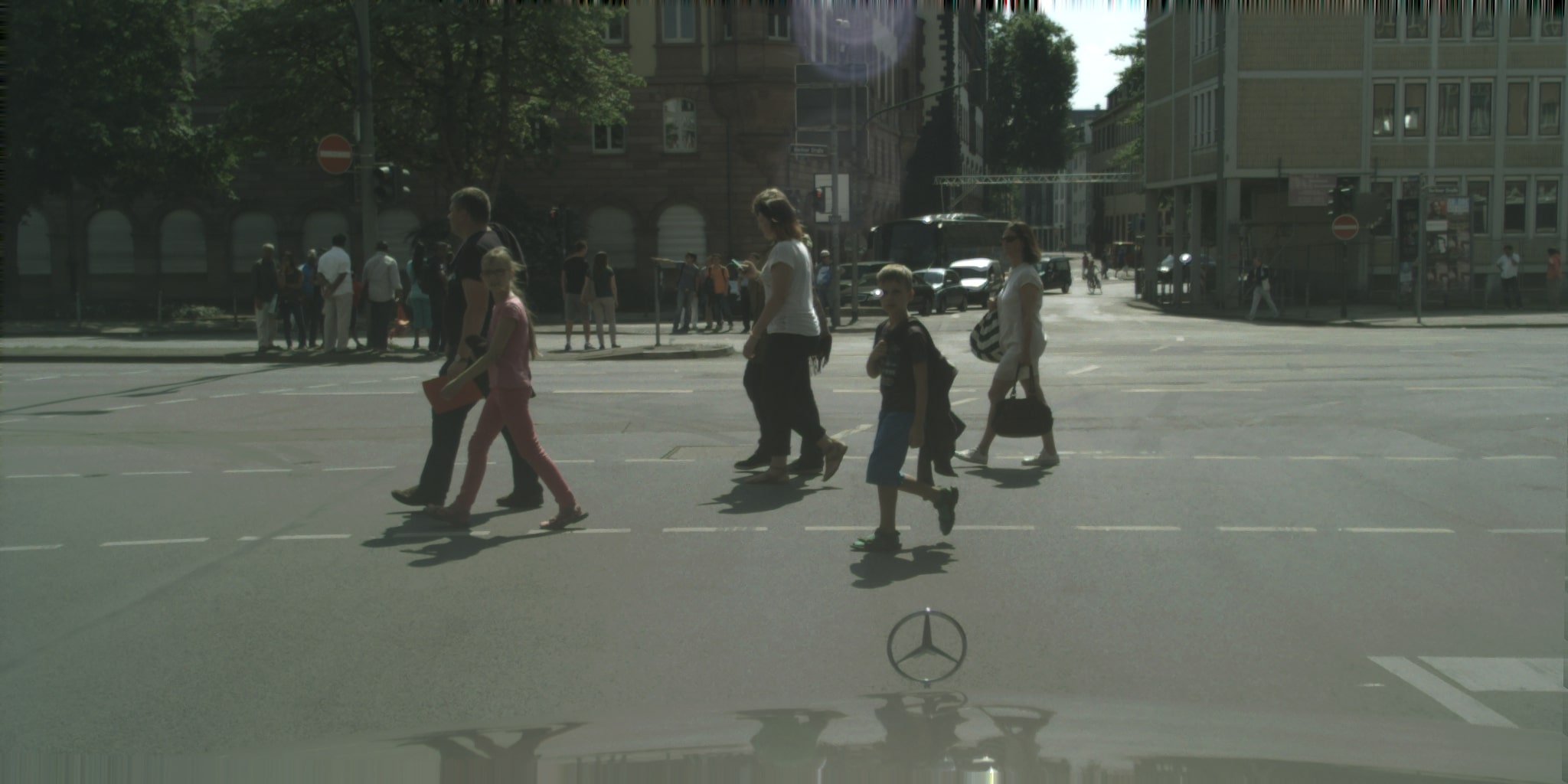}
\end{minipage}}
\subfigure[Source Only]{
\begin{minipage}[t]{0.24\linewidth}
\centering
\includegraphics[width=1.65in,height=0.83in]{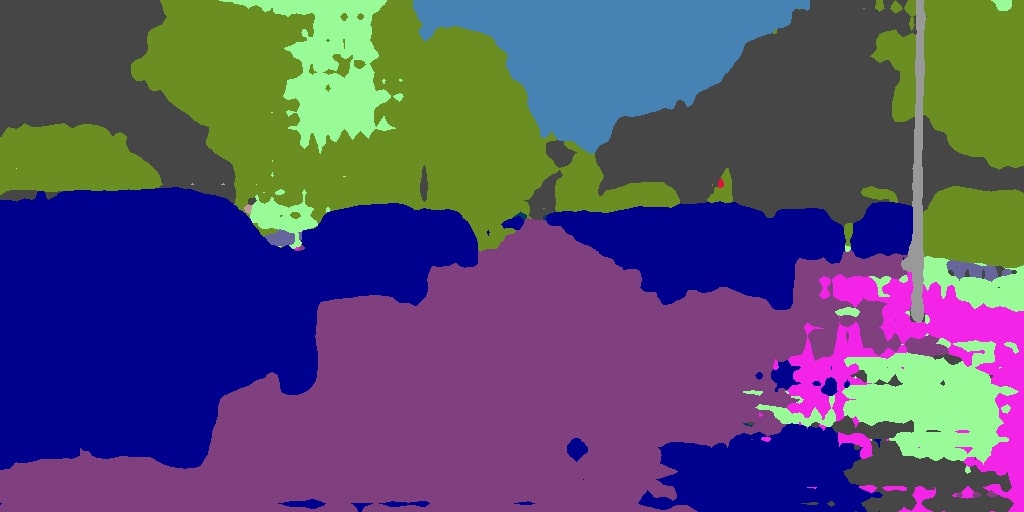}
\includegraphics[width=1.65in,height=0.83in]{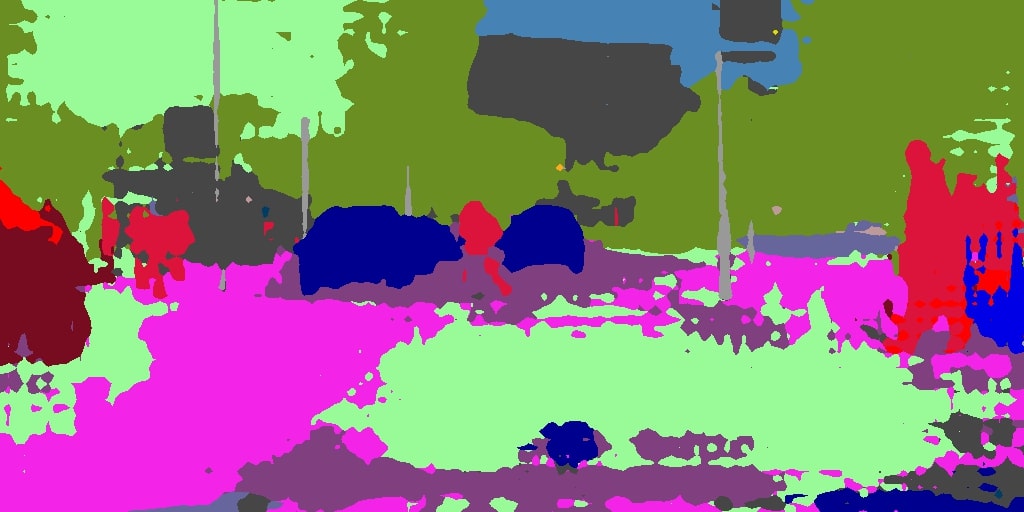}
\includegraphics[width=1.65in,height=0.83in]{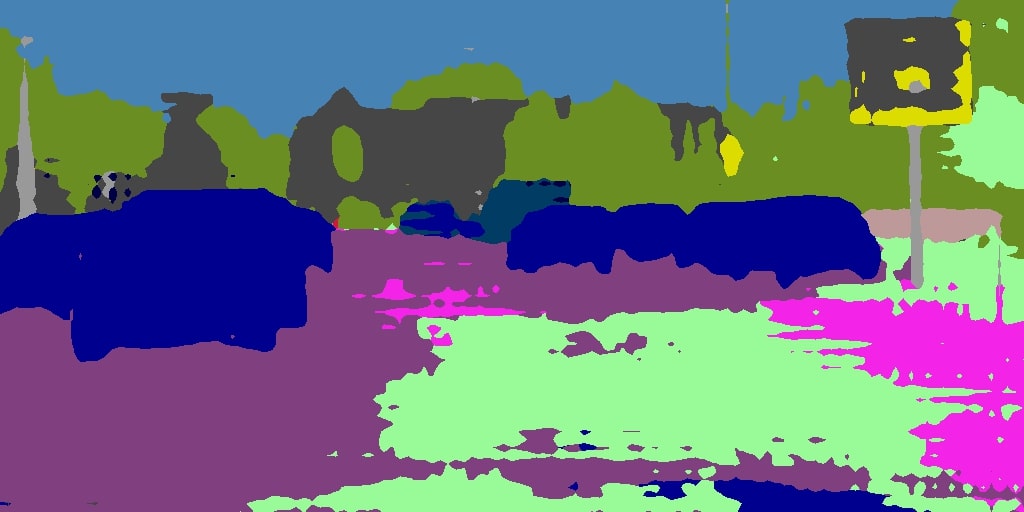}
\includegraphics[width=1.65in,height=0.83in]{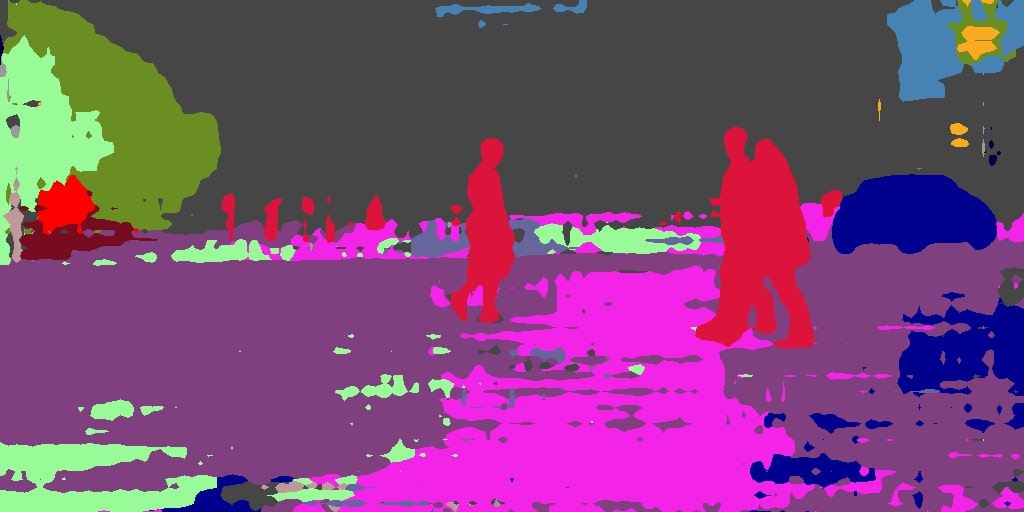}
\includegraphics[width=1.65in,height=0.83in]{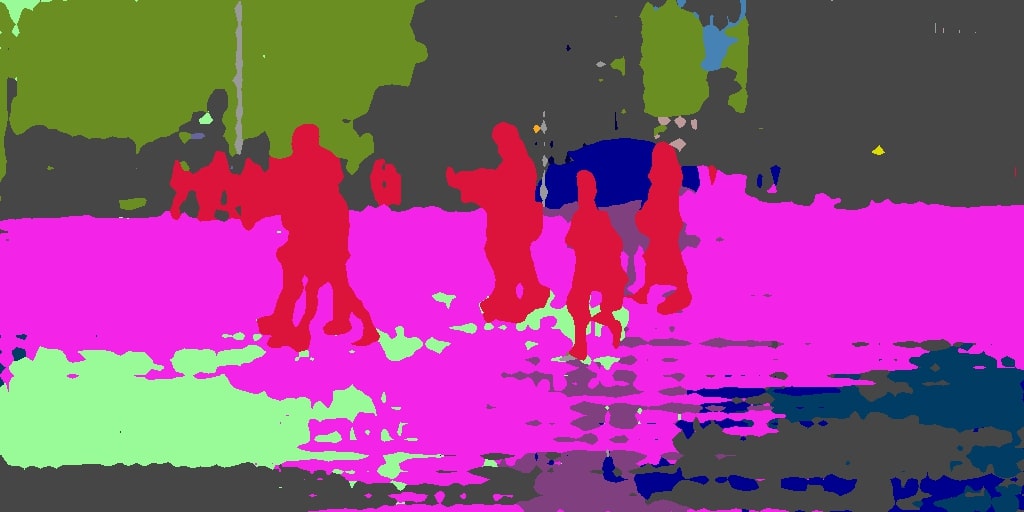}
\end{minipage}}
\subfigure[Ours]{
\begin{minipage}[t]{0.24\linewidth}
\centering
\includegraphics[width=1.65in,height=0.83in]{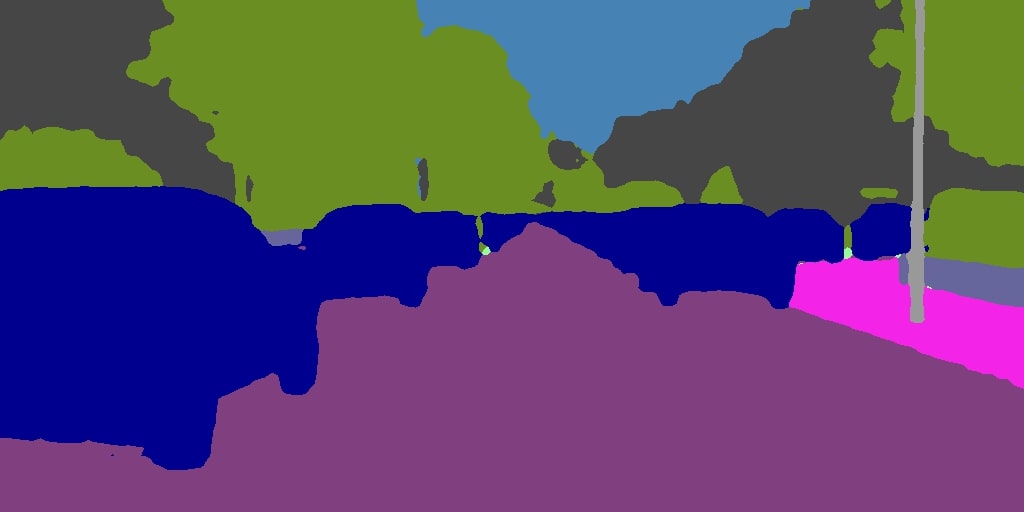}
\includegraphics[width=1.65in,height=0.83in]{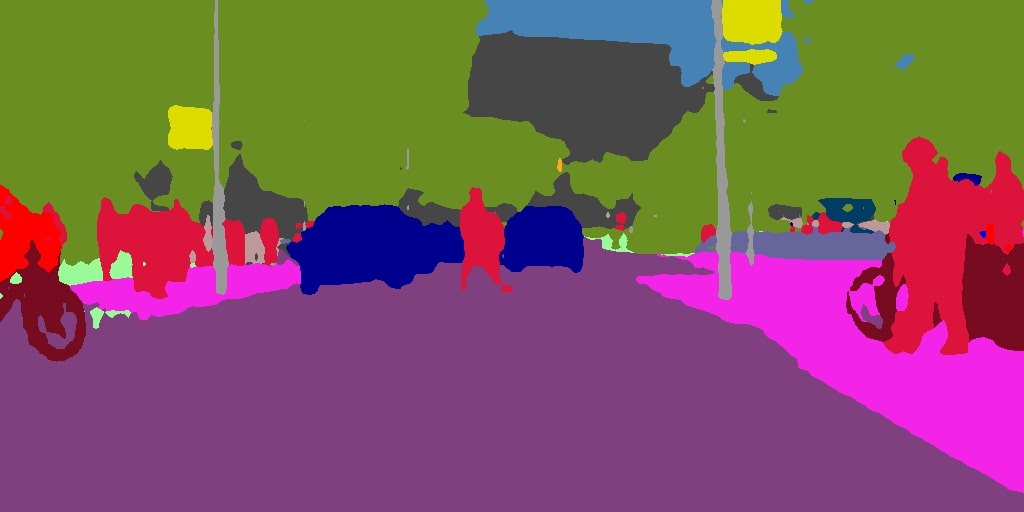}
\includegraphics[width=1.65in,height=0.83in]{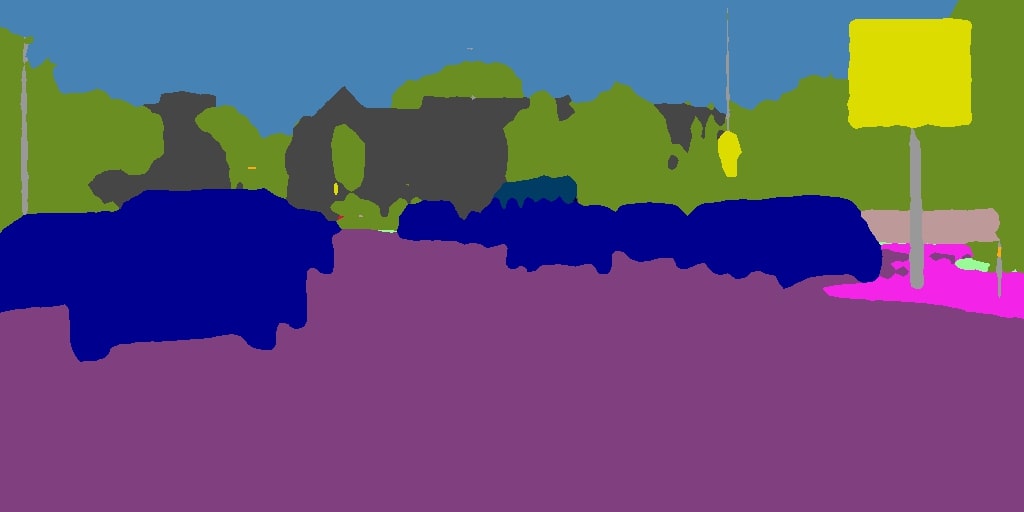}
\includegraphics[width=1.65in,height=0.83in]{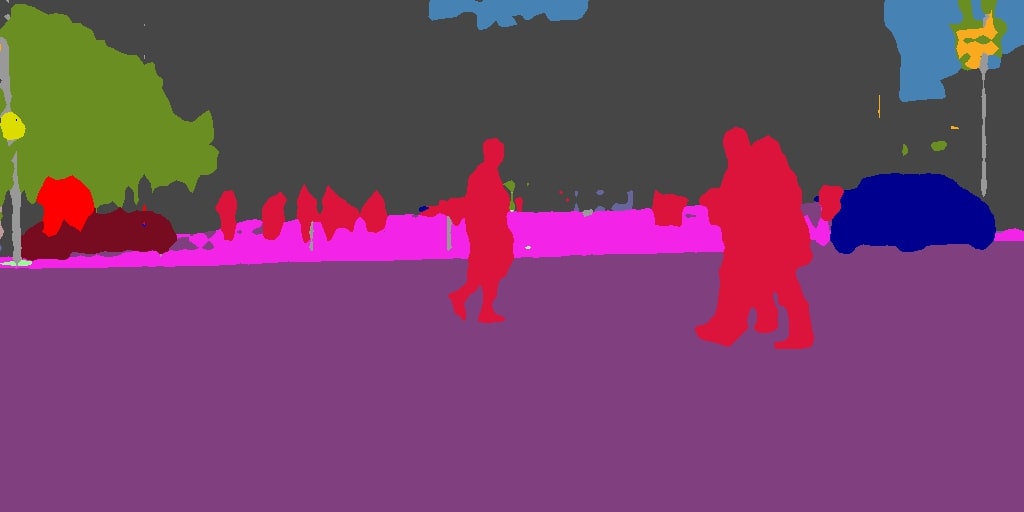}
\includegraphics[width=1.65in,height=0.83in]{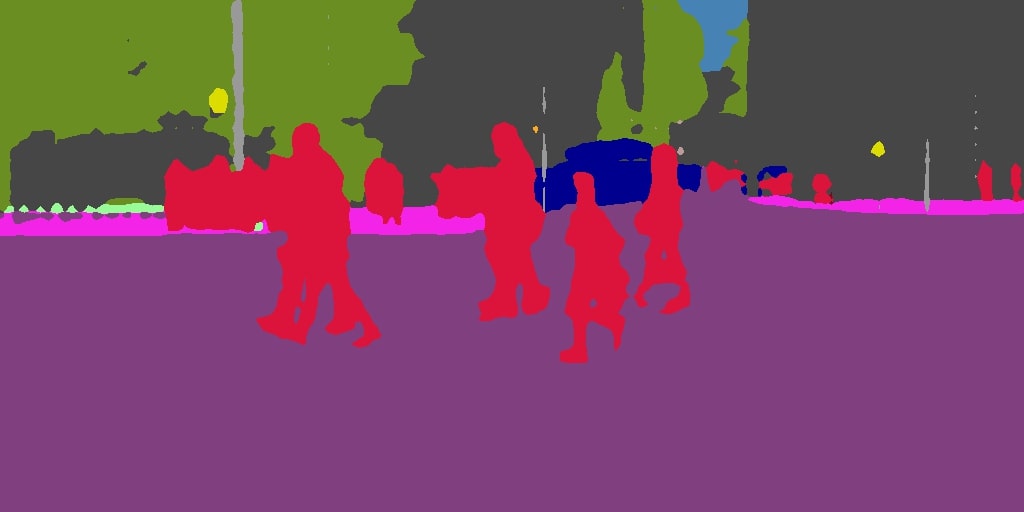}
\end{minipage}}
\subfigure[Ground Truth]{
\begin{minipage}[t]{0.24\linewidth}
\centering
\includegraphics[width=1.65in,height=0.83in]{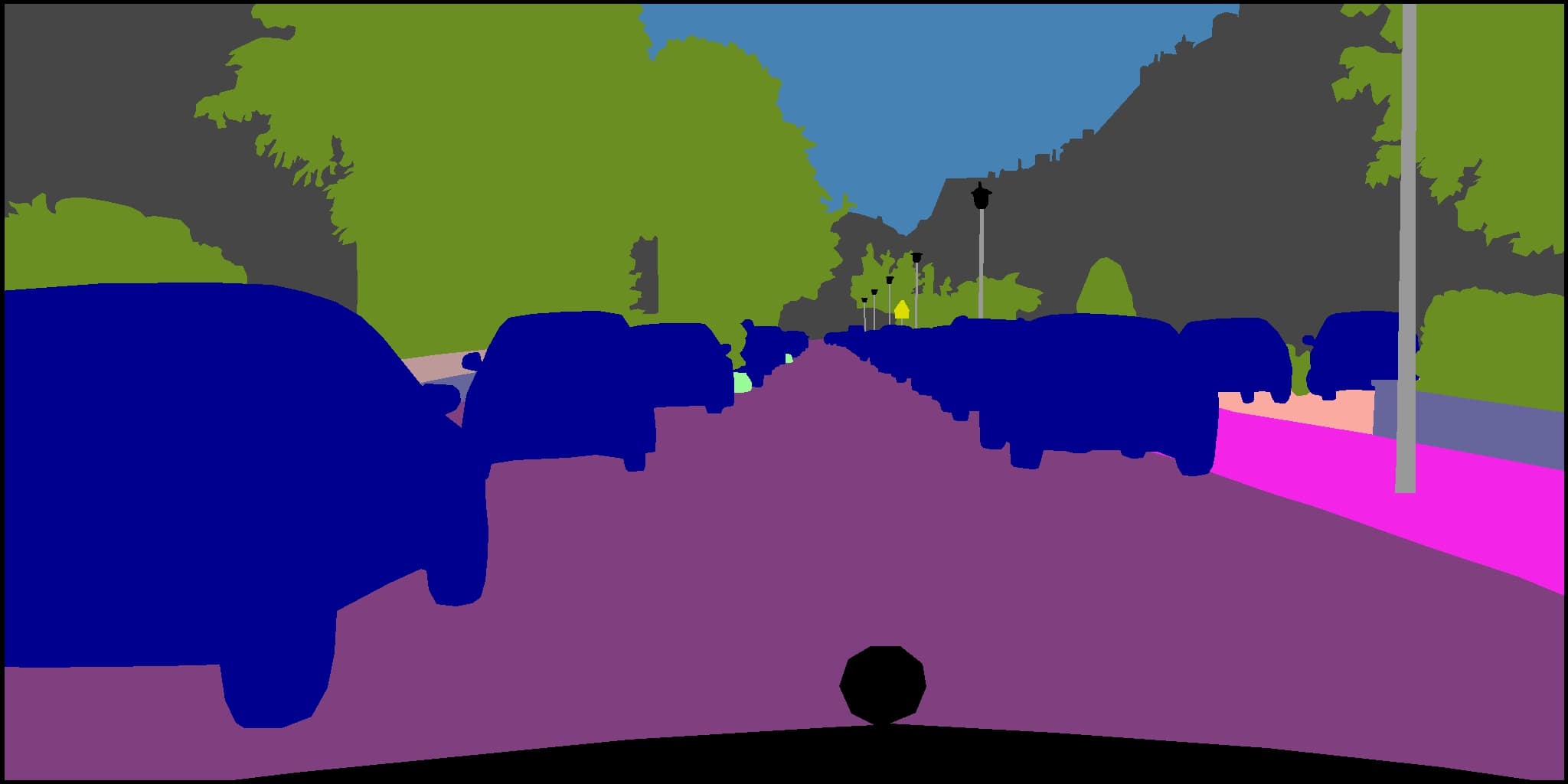}
\includegraphics[width=1.65in,height=0.83in]{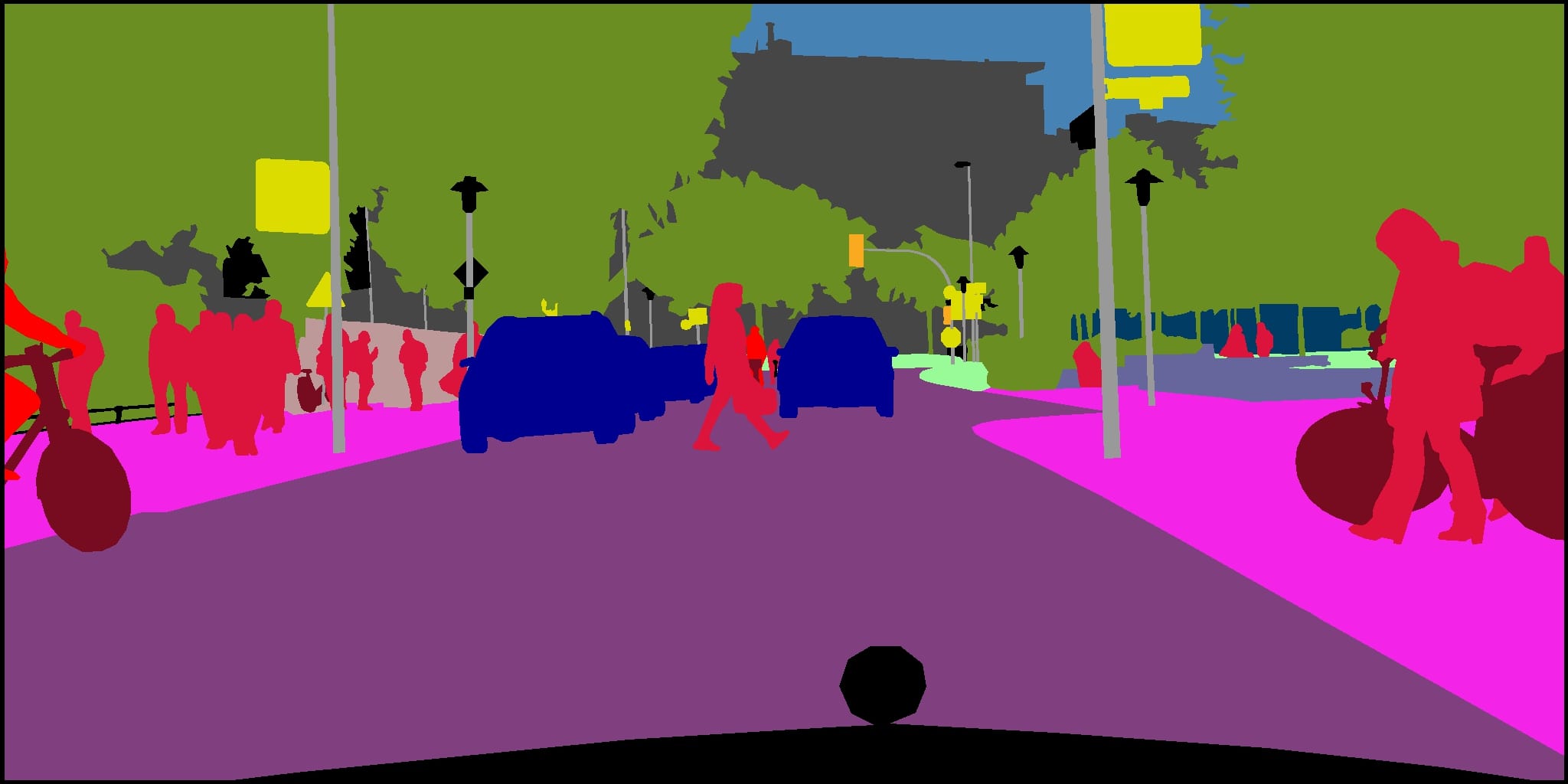}
\includegraphics[width=1.65in,height=0.83in]{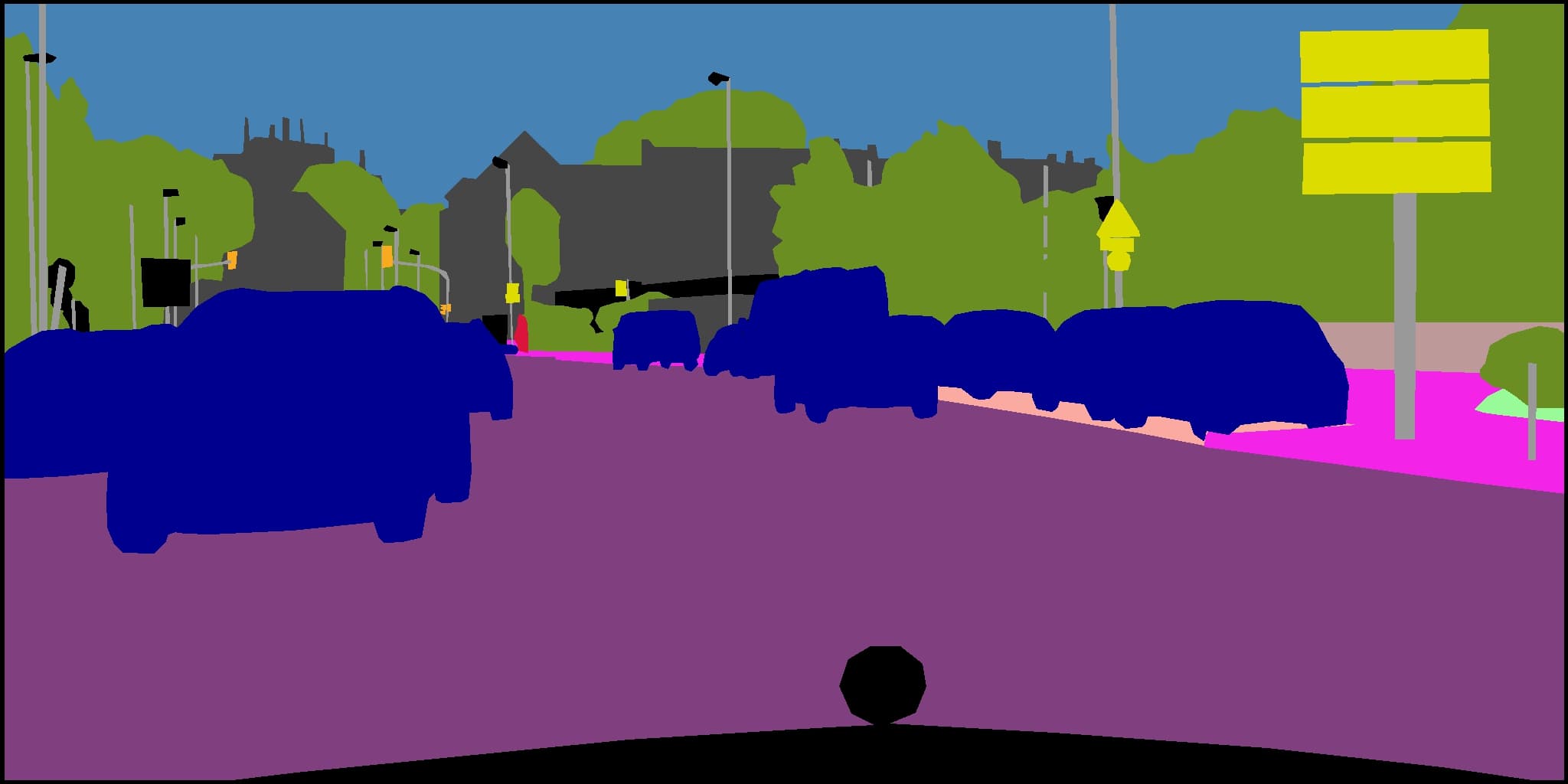}
\includegraphics[width=1.65in,height=0.83in]{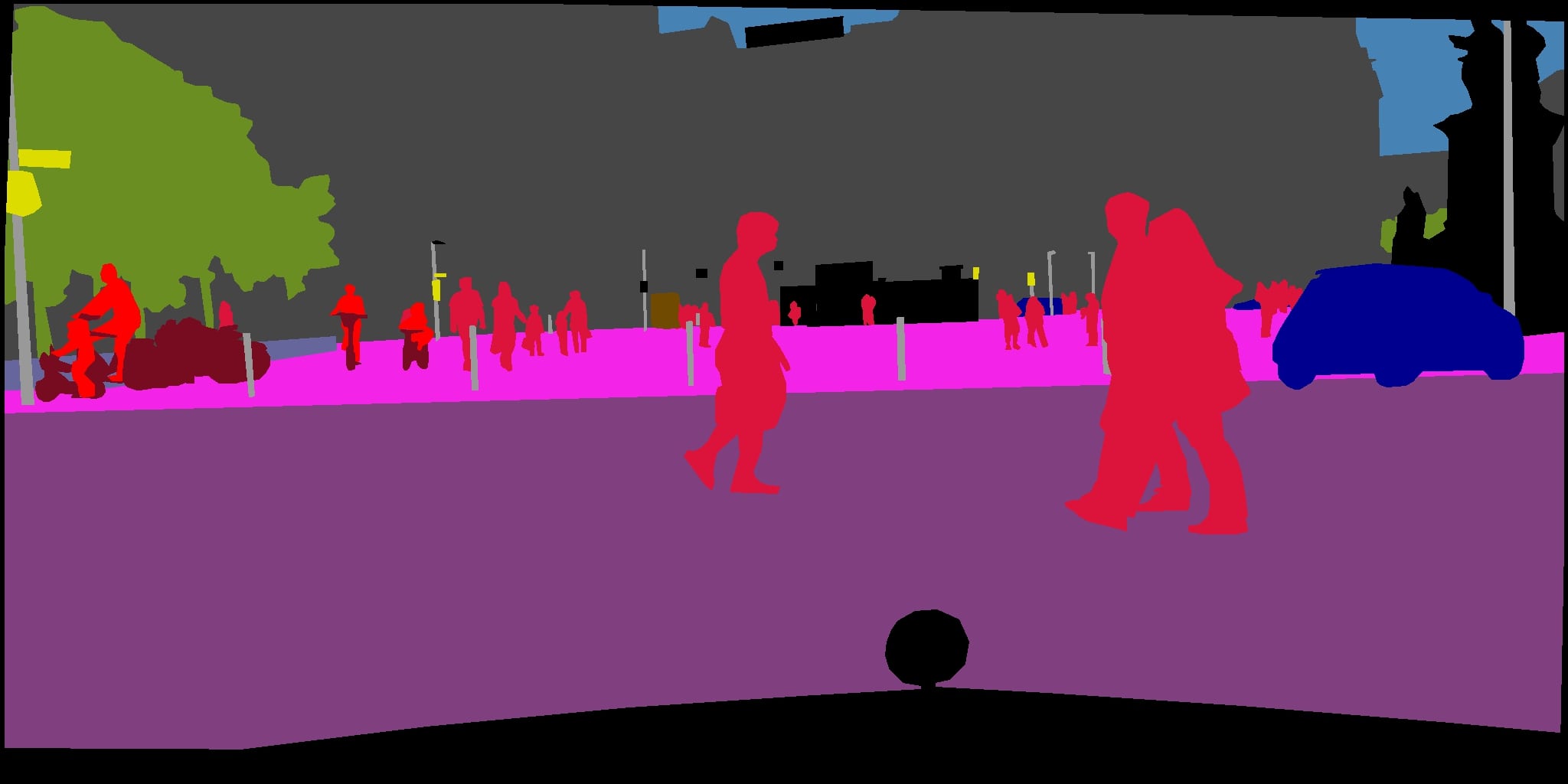}
\includegraphics[width=1.65in,height=0.83in]{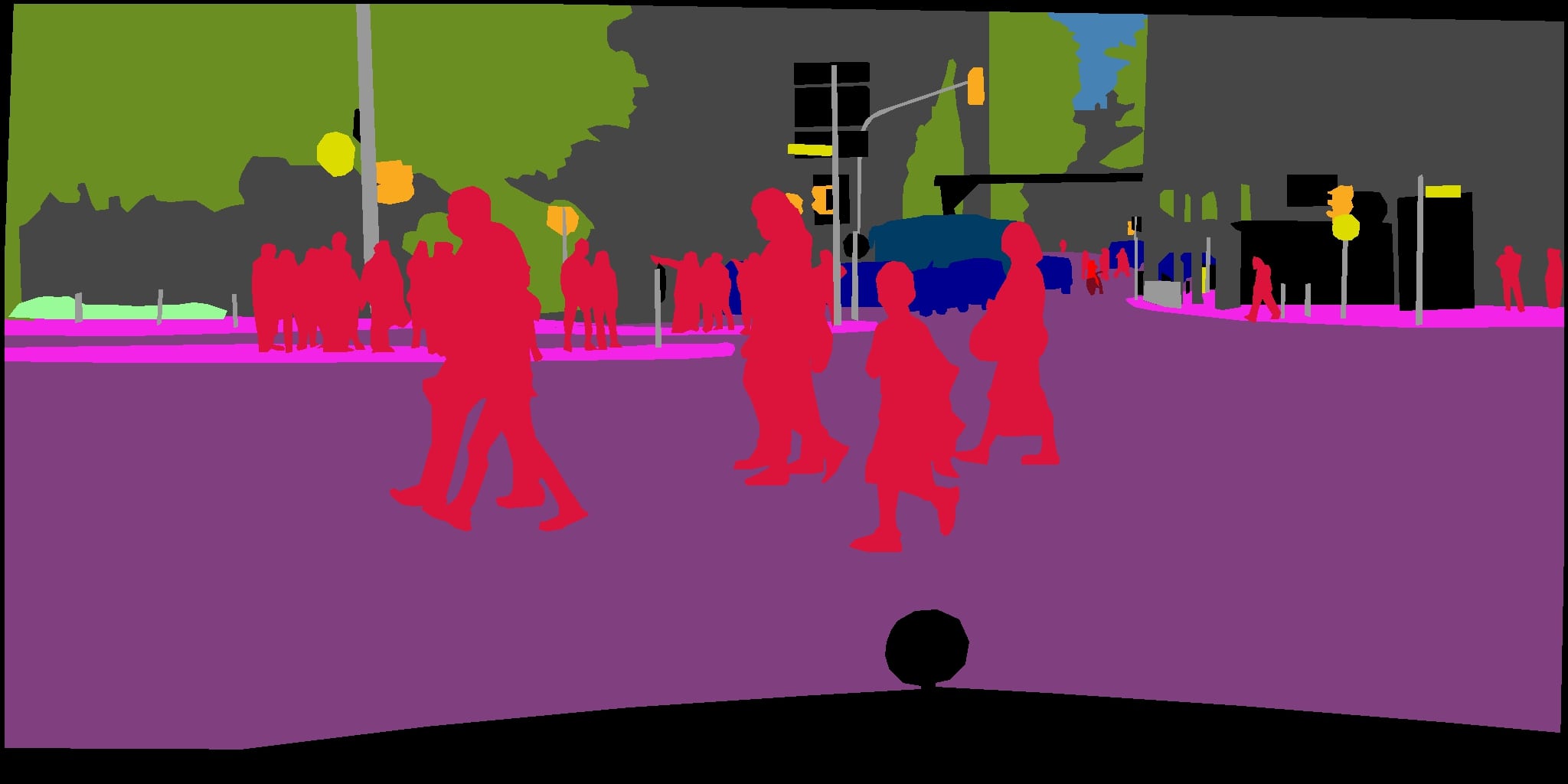}
\end{minipage}}
\centering
\caption{Some visual segmentation results for the GTA5 to Cityscapes task.}
\label{visual}
\end{figure*}

Table \ref{gta2city} shows the results of different methods for the GTA5 to Cityscapes task over 19 common classes. Our DSP model achieves the best mIoU score of 55.0\%, significantly outperforming state-of-the-art methods by large margins from 2.8\% to 11.4\%. Besides, our method shows its effectiveness in predicting long-tail classes, 
Moreover, DSP outperforms the source-only segmentation model by 18.4\% mIoU, showing a good cross-domain generalization ability. As for another SYNTHIA to Cityscapes task, we report the mIoU results of both 13 and 16 classes in Table \ref{syn2city}. As can be seen, DSP achieves the best performance in terms of mIoU of 16 and 13 classes, $i.e.$, 51.0\% mIoU and 59.9\% mIoU*, respectively. It also outperforms all existing methods, 
and achieves a significant improvement over the source-only model by 17.5\% and 21.3 \% over 16 and 13 classes in terms of mIoU. 

\subsection{Parameter Analysis and Ablation Study}
\label{subsec:ablationstudy}
To investigate the impact of different components in our DSP model, we conduct an ablation study on the GTA5 to Cityscapes setting as shown in Table \ref{ablation}. ``Source Only'' denotes the model without domain adaptation. ``Mean Teacher'' (MT) denotes the vanilla mean teacher framework. ``Single Paste'' denotes the pasting strategy proposed in DACS \cite{dacs} that only pasting source image patches to the target domain images. ``Dual Paste'' denotes the dual paste strategy based on hard-paste while ``Dual Soft-Paste'' denotes the proposed dual paste strategy based on soft-paste. ``Feature Alignment'' denotes the DSP-induced feature-level alignment. As can be seen, the source only model obtains 36.6 mIoU on the target domain. After using the mean teacher framework, a gain of 5.7\% mIoU can be observed. The single paste strategy (DACS) brings another 9.8\% mIoU improvement, 15.5\% mIoU in total. By contrast, our dual soft-paste strategy achieves a gain of 17.9\% mIoU over the source only baseline model. Compared with hard-paste, the proposed soft-paste strategy is more effective. After using the proposed DSP-induced feature-level alignment, our DSP model improves the baseline model by a significant margin, $i.e.$, 18.4\% mIoU. 

The training losses and mIoU results of DACS \cite{dacs}, $i.e.$, single hard-paste, Dual Hard-Paste (a variant of the proposed DSP based on hard-paste), and the proposed DSP on the GTA5 to Cityscapes setting are plotted in Figure \ref{loss_miou}. In the early training phase, DACS suffers from the large domain gap and may produce incorrect predictions, especially for those long-tail classes. Consequently, these incorrect predictions may mislead the adaptation process, leading to a slow convergence speed and limited performance. By contrast, our model adopts a dual soft-paste strategy and a long-tail class first sampling strategy to create intermediate domains having smaller domain gaps, thereby facilitating the domain adaptation. In addition, it can be seen that the dual paste strategy contributes to the faster convergence speed while the soft-paste strategy matters for better cross-domain generalization performance. 

\begin{table}[htb]
    \centering
\begin{tabular}{l|c|c}

\hline

\multicolumn{3}{c}{GTAV $\to$ Cityscapes}\\
\hline
   Methods  & mIoU (\%)&Gain(\%) \\
   \hline
   \hline
   Source Only& 36.6&-\\
  +Mean Teacher (MT) &42.3&5.7\\
  +MT + Single Paste&52.1&15.5\\
  \hline
   +MT + Dual Paste& 53.6&17.0\\
   +MT + Dual Soft-Paste&54.5 &17.9\\
   +MT + DSP + Feature Alignment&55.0 &18.4\\
   \hline
\end{tabular}
    \caption{Ablation study of the proposed DSP model.}
    \label{ablation}
\end{table}

Table \ref{opacity} shows the results of different hyper-parameter settings of the opacity $\beta$. When $\beta=0.8$, the model achieves the best performance, $i.e.$, 55.0\% mIoU. When $\beta=0$, the model is the vanilla mean teacher model, which only obtains 42.3\% mIoU. And when $\beta = 1$, the model becomes the mean teacher model using the dual hard-paste strategy, obtaining a better mIoU of 53.6\%. Besides, when $\beta$ is less than 0.7, the weight of the pasted source image patch is too small, which may result in inaccurate predictions of target images, especially in the early training phase, thereby affecting the final performance. 

\begin{table}[htb]
    \centering
    \begin{tabular}{c|c|c|c|c|c|c|c}
    \hline
    \multicolumn{8}{c}{GTAV $\to$ Cityscapes}\\
    \hline
    \hline
      $\beta$  & 0& 0.5&0.6&0.7&0.8&0.9&1\\
      \hline
      mIoU   & 42.3&51.0&52.2&54.5&\textbf{55.0}&54.9&53.6\\
      \hline
    \end{tabular}
    \caption{Hyper-parameter study of opacity $\beta$.}
    \label{opacity}
\end{table}

Table \ref{long-tail} shows the results of using different numbers of long-tailed classes $k$ during pasting. As can be seen, the performance peaks at $k=2$. When $k=0$, the performance drops by a margin of 1.8\% mIoU, implying the proposed long-tail class first sampling strategy matters for mitigating the class imbalance issue. Besides, when $k$ becomes larger, the sampled patches from different images may overlap each other and also result in inconsistent spatial layouts in the pasted patch, which will affect the performance.

\begin{table}[htb]
    \centering
    \begin{tabular}{c|c|c|c|c}
    \hline
    \multicolumn{5}{c}{GTAV $\to$ Cityscapes}\\
    \hline
    \hline
      long-tail classes to choose  & 0& 1&2&3\\
      \hline
      mIoU   &53.2&54.3&\textbf{55.0}&53.7\\
      \hline
    \end{tabular}
    \caption{Hyper-parameter study of the number of long-tailed classes $k$.}
    \label{long-tail}
\end{table}

In Figure \ref{visual}, we present some visual segmentation results of the source only model and our DSP model. As can be seen, the source only model has limited cross-domain generalization ability without any domain adaptation. There are large areas of incorrect predictions, such as road and trees. By contrast, our DSP model shows a fairly good cross-domain generalization performance.

\section{Conclusion}
In this paper, we investigate the unsupervised domain adaptive semantic segmentation problem from the perspective of image manipulation. Specifically, we propose a novel Dual Soft-Paste (DSP) method to create new intermediate domains with smaller domain gaps. Based on the mean teacher framework, DSP-induced output-level alignment and feature-level alignment are performed, which help to learn domain-invariant features. Besides, the long-tail class first sampling strategy used in DSP shows its effectiveness in addressing the class-imbalance issue. Experiments on two challenging benchmarks demonstrate the superiority of DSP over state-of-the-art methods. In the future, we plan to investigate the impact of DSP in other domain adaptation frameworks as well as develop an adaptive sampling strategy using reinforcement learning, which can actively sample both normal and long-tail classes during training. 
\section{ACKNOWLEDGMENTS}
This work was supported by the Science and Technology Major Project of Hubei Province (Next-Generation AI Technologies) under Grant 2019AEA170 and the Fundamental Research Funds for the Central Universities under Grant 2042021kf0196. The numerical calculations in this paper had been supported by the supercomputing system in the Supercomputing Center of Wuhan University.

\bibliographystyle{ACM-Reference-Format}
\bibliography{refer}

\end{document}